\documentclass{article} 
\usepackage[preprint]{colm2026_conference}

\usepackage{microtype}
\usepackage{hyperref}
\usepackage{url}
\usepackage{booktabs}

\definecolor{darkblue}{rgb}{0, 0, 0.5}
\hypersetup{colorlinks=true, citecolor=darkblue, linkcolor=darkblue, urlcolor=darkblue}

\usepackage{lineno}

\usepackage{amsmath}
\usepackage{amssymb}
\usepackage{mathtools}
\usepackage{amsthm}
\usepackage{algorithm}
\usepackage{algorithmic}
\usepackage{subcaption}

\title{Beyond Perplexity: Character Distribution Signatures and the MDTA Benchmark for AI Text Detection}

\author{Priyadarshan Narayanasamy$^{1}$\thanks{Equal contribution.} \\
\texttt{nspd@umd.edu}
\And
Swastik Agrawal$^{1}$\footnotemark[1] \\
\texttt{swastik3@umd.edu}
\And
Klint Faber$^1$ \\
\texttt{kfaber@umd.edu}
\And
Fardina Fathmiul Alam$^{1}$\thanks{Corresponding author.} \\
\texttt{fardina@umd.edu} \\[0.5em]
$^1$ Department of Computer Science, University of Maryland, College Park
}

\begin{document}

\ifcolmsubmission
\linenumbers
\fi

\maketitle

\begin{abstract}

Training-free AI text detection methods primarily rely on model log-probabilities, achieving strong performance through approaches like Binoculars and DNA-DetectLLM. However, these methods face a fundamental ceiling as models are optimized through RLHF to produce human-like probability distributions. We introduce an alternative detection signal based on character distribution signatures. We provide theoretical foundations showing that AI models, trained on massive domain-balanced corpora, approximate global character patterns while humans exhibit domain-specialized distributions, creating a ``Wall of Separation" where human-AI divergence significantly exceeds AI-AI divergence. To enable systematic evaluation, we construct the Models-Domains-Temperatures-Adversarials (MDTA) benchmark comprising 642,274 prompt-aligned samples across 4 models, 5 domains, 3 temperature settings, and 3 adversarial strategies, substantially expanding the HC3 dataset with modern model responses, temperature variation, and adversarial augmentation. We introduce the Letter Distribution Score (LD-Score), demonstrating low correlation ($r = 0.08 \text{--} 0.13$) with perplexity methods. When integrated with DNA-DetectLLM, Binoculars and FastDetectGPT via a non-linear classifier, LD-Score yields consistent improvements in AUROC and F1, with particularly pronounced gains in specialized domains where vocabulary constraints amplify the detection signal. The MDTA dataset can be accessed at: \url{https://huggingface.co/datasets/nsp909/MDTA}

\end{abstract}

\section{Introduction}
\label{sec:Introduction}

Large Language Models (LLMs) now generate content across countless domains, but distinguishing AI text from human writing has become critical in high-stakes applications. In academia, detection prevents plagiarism in student assignments and protects research integrity which is a pressing concern given that AI-generated manuscripts with fabricated citations have already infiltrated peer review at major conferences \citep{nichols2025hallucinated}. Detection also guards against hallucinated information in legal documents and medical advice, where factual errors can have serious consequences.

Current state-of-the-art perplexity-based approaches, such as Binoculars~\citep{bino} and DNA-DetectLLM~\citep{zhu2025}, achieve strong detection performance by analyzing model log-probabilities. However, these methods face a fundamental ceiling as models are optimized to mimic human likelihood distributions, motivating complementary approaches that capture fundamental properties of text beyond perplexity.

We introduce an orthogonal detection signal based on letter distribution signatures. LLMs operate with comprehensive word probability distributions derived from extensive vocabularies and massive training corpora, while individual human writers exhibit specialized and skewed distributions with domain-specific patterns. This fundamental asymmetry causes LLMs to approximate global letter level statistical patterns, while humans deviate significantly through constrained vocabularies and stylistic preferences, creating a detectable signature at the letter level.

We introduce the Letter Distribution Score (LD-Score), an interpretable metric that quantifies letter distribution divergence. We provide mathematical foundations for why letter distributions differ systematically between human and AI text, then empirically validate the ``Wall of Separation'', revealing that letter distribution divergence between human and AI text noticeably exceeds divergence between different AI models.

Existing benchmark datasets present critical limitations: they either lack model-temperature diversity, domain coverage, prompt alignment, adversarial augmentation, or sufficient scale for robust evaluation. We contribute the Models-Domains-Temperatures-Adversarial (MDTA) benchmark, consisting of 642,274 prompt-aligned samples across 4 models, 5 domains, 3 temperature settings, and 3 adversarial strategies, substantially expanding the HC3 dataset~\citep{guo-etal-2023-hc3} with modern model responses, temperature variation, and targeted adversarial attacks including lipogrammatic constraints designed to challenge character-level detection methods.

We demonstrate that LD-Score provides an orthogonal detection signal to perplexity-based methods, exhibiting low correlation ($r = 0.08 \text{--} 0.13$) with existing approaches. Integration with state-of-the-art approaches (Binoculars, DNA-DetectLLM) yields consistent improvements in AUROC, and F1 scores, with the most notable gains observed in technical domains such as finance and medicine, where restricted vocabularies produce stronger character-level separation between human and AI text. Our approach requires no internal model access, providing a practical and effective complement to existing detection methods.

\section{Related Work}
\label{sec:Background}

AI-generated text detection has been an active research area since the early days of GPT-2, when ~\citet{solaiman} developed the first major neural detector, a fine-tuned RoBERTa model. Since then, numerous deep learning approaches have emerged, with recent work reporting very high accuracies (95-99\%) using fine-tuned BERT~\citep{wang2024}, transformer architectures combined with linguistic features, and hybrid methods leveraging Bi-LSTM with attention mechanisms~\citep{blake}. Despite these impressive results in controlled settings, deep learning detectors face significant limitations. They struggle to generalize across different language models and domains, and are vulnerable to simple adversarial attacks~\citep{sandy}, with detectors trained on older models failing to detect outputs from newer models within the same generation or family. These limitations motivate the need for detection approaches that rely on more fundamental properties, both language-based and LLM-based, rather than learned model-specific patterns.

DetectGPT~\citep{mitchell2023detectgptzeroshotmachinegeneratedtext} established an important foundation for training-free detection by showing that text sampled from an LLM tends to occupy negative curvature regions of the model's log-probability function. Binoculars~\citep{bino} introduced a contrastive framework that uses two models of different sizes to measure log-probability divergence, while DNA-DetectLLM~\citep{zhu2025} refined detection through probability perturbations across different model states and sampling strategies. More recently, BISCOPE~\citep{biscope} proposed a related but distinct logit-based framework that uses surrogate LLMs to extract bidirectional token-level features from forward and backward cross-entropy, rather than relying solely on standard next-token probability criteria.  While these methods achieve strong empirical performance, they face fundamental limitations. As models are increasingly optimized through RLHF and constitutional AI to produce more human-like probability distributions, the probability gap narrows, creating a detection ceiling. Moreover, many of these approaches rely on access to model log-probabilities or logits, limiting their applicability in black-box settings. These limitations motivate the need for orthogonal detection signals that operate independently of model probability distributions and can augment existing methods.

Stylometric methods distinguish human from AI text by analyzing writing style features, such as \citet{kumarage} who applied this approach to detect AI-generated tweets in social media timelines, \citet{li-zhang} developed a comprehensive framework for Chinese social media with 34 features. These methods typically rely on features such as punctuation frequency, phraseology patterns, lexical complexity, sentence length statistics, linguistic diversity metrics, readability scores, and sentiment markers. However, these stylometric features represent surface-level characteristics rather than fundamental properties of text generation. They can be easily circumvented through fine-tuning, as models can be trained to mimic specific stylistic patterns once these features are identified by detectors.

$N$-gram frequency analysis represents a classical and popular approach to authorship attribution that has been applied to AI text detection. \citet{galle} demonstrated that repeated higher-order $n$-grams over-appear in machine-generated text, while \citet{yang2023} proposed Divergent $N$-Gram Analysis (DNA-GPT), a training-free detection strategy that analyzes differences between original and regenerated text portions through $n$-gram analysis. Despite being well-established, $n$-gram methods face severe limitations for AI text detection: higher-order $n$-grams grow prohibitively sparse and computationally expensive, models can be explicitly trained to avoid or mimic specific $n$-gram patterns, and word-level analysis inherits the same lexical dependencies and adversarial vulnerabilities as stylometric approaches.

Drawing inspiration from naturally occurring statistical patterns like Benford's law~\citep{benford} and Zipf's observations about language structure~\citep{zipf1949}, we hypothesize that letter-level distributions might encode detectable signatures based on the fundamental structure of language itself. Unlike word-level features that can be easily manipulated, letter distributions emerge from the aggregate effect of vocabulary selection and usage patterns, representing a more fundamental property of text generation. This intuition led us to develop the Letter Distribution Score as an orthogonal detection signal.

\section{Letter Distribution Signatures}
\label{sec:The Letter Distribution Score}

\subsection{Theoretical Foundation: Exposure Scale and Domain Diversity}
\label{sec:theory-foundation}

We formalize the fundamental asymmetry between human and AI text generation through their differential exposure to natural language and the resulting proximity to the global word probability distribution.

\textbf{The Global Word Distribution.} Let $P_{\text{global}}(w)$ denote the true word probability distribution across natural language - the population distribution aggregated over all contexts, domains, speakers, and time periods.

\textbf{Convergence Through Exposure.} When observing $N$ words from a source, we estimate its word probability distribution $P(w)$. We treat the empirical word distribution as the marginal probability of observing each word. By the Law of Large Numbers, the empirical probability distribution converges to the true distribution with approximation error bounded by~\citep{boucheron}:
\begin{equation}
\left\| P(w) - P_{\text{global}}(w) \right\| = O\left(\frac{1}{\sqrt{N}}\right)
\label{eq:empirical_distr_error}
\end{equation}
Larger exposure ($N$) yields better approximation, but only when samples are drawn proportionally from the target distribution. 

\textbf{AI Models: Massive, Domain-Balanced Exposure.} Even the smallest of modern training corpora contain approximately 1.0 trillion tokens~\citep{wanjuan-cc, shen2024slimpajama}. Using standard subword tokenization where one token $\approx$ 0.75 words, this yields $N_{\text{AI}} \approx 750$ billion words. These corpora span diverse domains: web text, Wikipedia, books, scientific papers, code, and conversational data. Under proportional sampling from $P_{\text{global}}$:
\begin{equation}
\left\| P_{\text{AI}}(w) - P_{\text{global}}(w) \right\| \approx \frac{1}{\sqrt{750 \times 10^9}} \approx 1.15 \times 10^{-6}
\label{eq:ai_distr_error}
\end{equation}

\textbf{Humans: Limited, Domain-Specialized Exposure.} Average adult reading speed is 238 words per minute~\citep{brysbaert2019}. Even assuming 8 hours daily (480 minutes) for 40 years:
\begin{equation}
N_{\text{human}} \approx 238 \times 480 \times 365 \times 40 \approx 1.67 \text{ billion words}
\label{eq:human_words_read_40_years}
\end{equation}
This represents a 300-fold difference compared to AI training data. Under proportional sampling:
\begin{equation}
\left\| P_{\text{human}}(w) - P_{\text{global}}(w) \right\| \approx \frac{1}{\sqrt{2.5 \times 10^9}} \approx 2.0 \times 10^{-5}
\label{eq:human_distr_error}
\end{equation}
approximately 17 times larger than AI model error.

\textbf{The Domain Specialization Bias.} The above assumes humans sample proportionally from $P_{\text{global}}$ -- an assumption that fails in practice. Humans exhibit strong domain specialization: medical professionals consume medical literature, engineers read technical documentation, and academics focus on scholarly work. Thus humans sample from domain-specific distributions $P_{\text{domain}} \neq P_{\text{global}}$. 

Total divergence decomposes as:
\begin{equation}
\begin{aligned}
\| P_{\text{human}}(w) - P_{\text{global}}(w) \| \approx\; 
&\underbrace{\frac{1}{\sqrt{N_{\text{human}}}}}_{\text{statistical}} 
+ \underbrace{\| P_{\text{domain}}(w) - P_{\text{global}}(w) \|}_{\text{domain bias}}
\end{aligned}
\label{eq:total_divergence}
\end{equation}
The domain bias is structural and persists regardless of reading volume. (We use ``domain" flexibly to refer to any coherent grouping of text)


\textbf{The Clustering Inequality.} State-of-the-art models (GPT-5, Claude 4.5, Gemini 3.0) are increasingly trained on overlapping corpora - primarily Common Crawl, Wikipedia, books, and scientific papers. This shared training causes them to approximate $P_{\text{global}}$ nearly identically:
\begin{equation}
P_{\text{GPT-5}}(w) \approx P_{\text{global}}(w) + \epsilon, \quad P_{\text{Claude}}(w) \approx P_{\text{global}}(w) + \epsilon'
\label{eq:ai_global_approximation}
\end{equation}
where $\epsilon \approx \epsilon'$ due to correlated data. This yields:
\begin{equation}
\boxed{
\max_{i,j \in \text{AI}} D(P_i, P_j) <
\min_{\substack{h \in \text{human}\\ a \in \text{AI}}} D(P_h, P_a)
}
\label{eq:ai_vs_human_inequality}
\end{equation}
where $D(\cdot,\cdot)$ denotes distributional divergence. This predicts that AI models form a tight cluster separated from human text: the ``Wall of Separation''.

\subsection{From Word Distributions to letter Distributions}
\label{sec:word-to-letter}

The above analysis establishes that $P_{\text{human}}(w)$ deviates from $P_{\text{global}}(w)$ due to domain specialization and limited exposure, while $P_{\text{AI}}(w) \approx P_{\text{global}}(w)$ due to massive and domain-balanced training. This divergence explains why word-level $n$-gram methods show empirical success. However, as discussed in Section~\ref{sec:Background}, word-level features face critical limitations: extreme sparsity, signal dilution across high-dimensional space, and vulnerability to adversarial manipulation.

We address these limitations by projecting word distributions to letter-level statistics. Define the letter distribution as:
\begin{equation}
P(\ell) = \sum_{w \in \mathcal{V}} P(w)\,\frac{L(w,\ell)}{|w|}
\label{eq:letter_distribution}
\end{equation}

where $\ell \in \{a, \ldots, z\}$, $L(w, \ell)$ counts occurrences of letter $\ell$ in word $w$, and $|w|$ is word length. This transformation achieves approximately 1,900 times dimensionality reduction (50,000 to 26) while amplifying discriminative signals through aggregation.

The aggregation effect is key: multiple words with similar discriminative patterns contribute to the same letters, causing weak word-level signals to accumulate into stronger letter-level signals concentrated in just 26 dimensions.


Because the word-to-letter transformation is linear in $P(w)$, the clustering inequality established earlier in this section transfers directly from word to letter space:
\begin{equation}
\boxed{
\max_{i,j \in \text{AI}} D(P_{i}^{\text{char}}, P_{j}^{\text{char}}) <
\min_{\substack{h \in \text{human}\\ a \in \text{AI}}} D(P_{h}^{\text{char}}, P_{a}^{\text{char}})
}
\label{eq:clustering_inequality}
\end{equation}
The ``Wall of Separation" persists at the letter level, preserving the discriminative structure while eliminating the sparsity, computational cost, and adversarial vulnerability inherent at the word-level.

\subsection{The Letter Distribution Score}
\label{lds}

We define the Letter Distribution Score (LD-Score) as a measure of letter-level distributional similarity between any two texts using a variation of Jensen-Shannon Distance (See Algorithm~\ref{alg:ld_score}). The score quantifies how closely two texts match in their letter usage patterns. Lower scores indicate greater similarity and suggest the texts likely originate from the same source (both human or both AI), while higher scores indicate divergence.

\begin{figure}[t]
    \centering
    \includegraphics[width=0.65\linewidth]{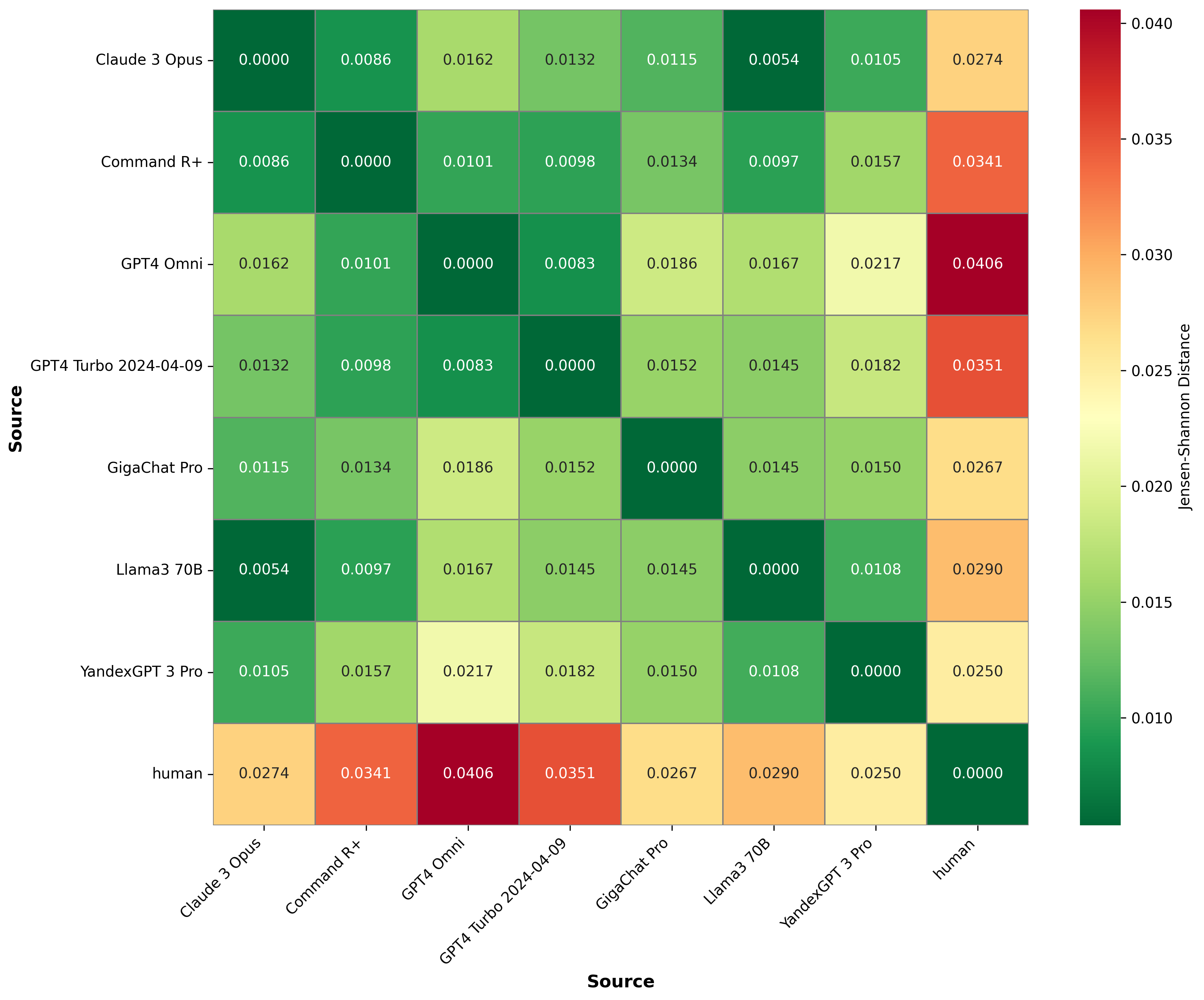}
   \caption{Pairwise LD-Scores between complete (A-Z) letter distributions in the Essay domain. The human row (bottom) exhibits systematically higher divergence from all AI models compared to AI-to-AI distances.}
    \label{fig:pairwise_essay}
\end{figure}

\subsection{Empirical Validation}
\label{sec:empirical_validation}

We validate our theoretical predictions using the Ghostbuster dataset~\citep{ghost}, analyzing LD-Score divergence across models and domains. For comprehensive domain-specific analysis and additional results, see Appendix~\ref{app:domain_analysis}.

\textbf{The Wall of Separation.} 
Figure~\ref{fig:pairwise_essay} presents pairwise LD-Scores in the essay domain. The matrix reveals the predicted two-scale structure: AI models cluster tightly with inter-model LD-Scores ranging from 0.0054 to 0.0217 (cool colors), while human text maintains consistently higher divergence from all AI models, with LD-Scores spanning 0.0250 to 0.0406 (warm colors).


Within the AI cluster, models with shared training backgrounds exhibit minimal LD-Score divergence: GPT-4 Omni and GPT-4 Turbo show LD-Score 0.0083, while Claude 3 Opus and Llama3 70B maintain 0.0054 - the smallest AI-AI LD-Score observed. This validates our theoretical prediction that overlapping training corpora induce correlated approximations to $P_{\text{global}}(w)$.

\begin{figure}[t]
    \centering
    \includegraphics[width=0.65\linewidth]{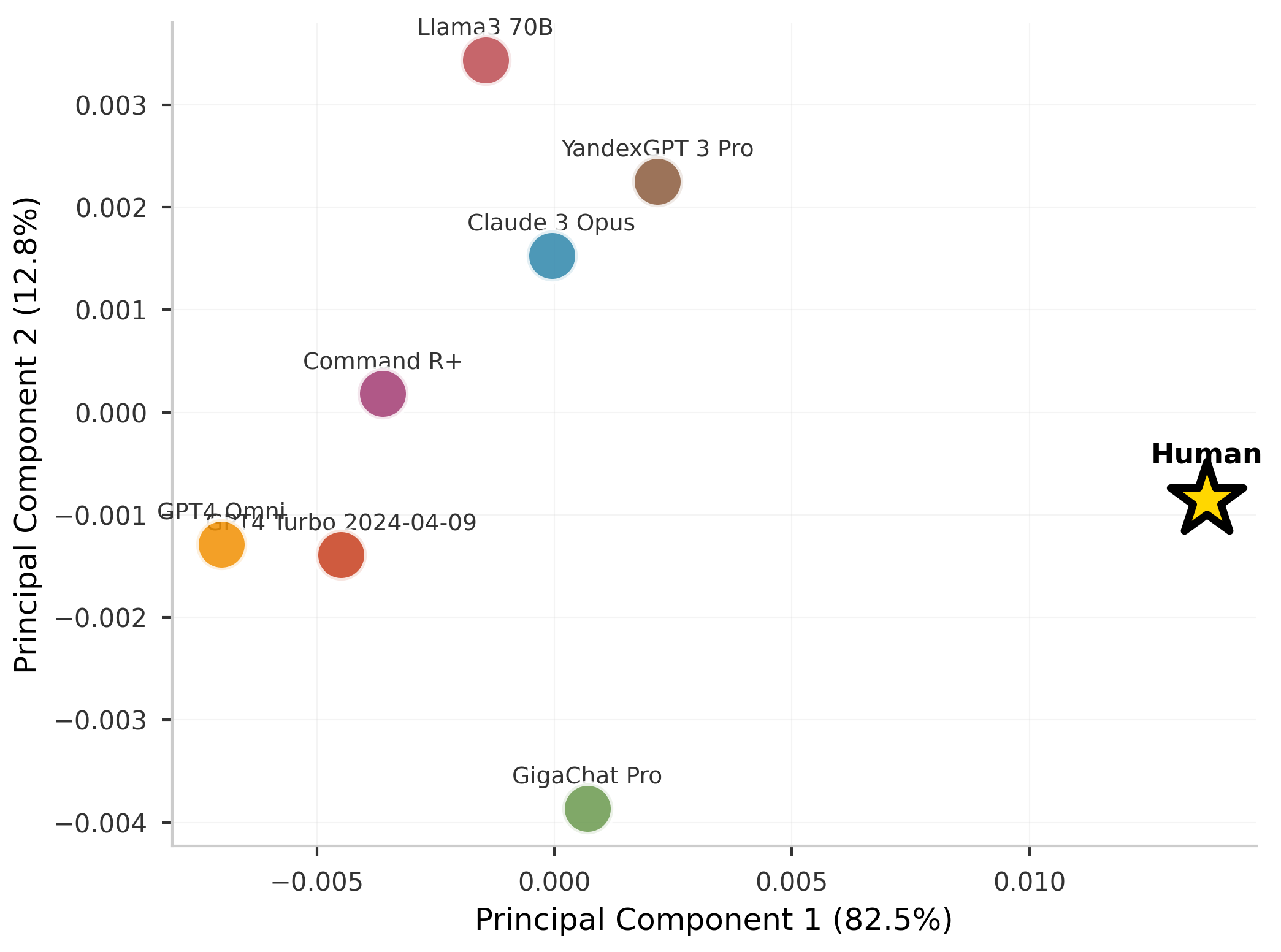}
   \caption{Principal Component Analysis (PCA) of letter probability distributions shows the human letter probability distribution is clearly distinct from all AI models, which cluster separately, indicating a consistent distributional shift in AI-generated text.}
    \label{fig:essay_pca_Pool}
\end{figure}

Figure~\ref{fig:essay_pca_Pool} provides a geometric visualization of the separation through PCA projection. The first two principal components capture 95.3\% of total variance (PC1: 82.5\%, PC2: 12.8\%), indicating that letter distribution differences concentrate in a low-dimensional subspace. The spatial arrangement reinforces the idea that the distance from human to the nearest AI model substantially exceeds the maximum distance between any two AI models, geometrically confirming the clustering inequality established in the heatmap analysis.




\begin{table}[t]
\centering
\begin{tabular}{l|ccccc}
\toprule
& DNA & Bino & Lexical & WD & LD \\
\midrule
DNA          & 1.00 & 0.86 & 0.02 & 0.17 & 0.13 \\
Bino   & 0.86 & 1.00 & -0.02 & 0.10 & 0.08 \\
Lexical      & 0.02 & -0.02 & 1.00 & 0.22 & 0.11 \\
WD-Score     & 0.17 & 0.10 & 0.22 & 1.00 & 0.42 \\
LD-Score     & 0.13 & 0.08 & 0.11 & 0.42 & 1.00 \\
\bottomrule
\end{tabular}
\caption{Correlation matrix between detection methods. LD-Score exhibits low correlation with perplexity-based methods (DNA-DetectLLM, Binoculars), indicating an orthogonal detection signal. In contrast, DNA and Binoculars are strongly correlated, reflecting shared likelihood-based modeling assumptions.}
\label{tab:orthogonality}
\end{table}

\textbf{Orthogonality to Existing Methods.}
To assess orthogonality with existing detection approaches, we compute Pearson correlations between detection signals. We compare: (1) DNA-DetectLLM score differences between AI and human text, (2) Binoculars score differences, (3) stylometric feature distance (punctuation frequency, sentence length, lexical diversity), (4) word-level RJSD (WD-Score), and (5) LD-Score. For the remainder of the paper, we refer to DNA-DetectLLM as \textbf{DNA} and Binoculars as \textbf{Bino}.

Table ~\ref{tab:orthogonality} presents the correlation matrix. As expected, DNA and Bino exhibit high correlation (r=0.86), reflecting their shared reliance on perplexity signals. LD-Score demonstrates low correlation with both perplexity-based methods (r=0.13 with DNA, r=0.08 with Bino), confirming near-orthogonality. 


While stylometric features appear highly orthogonal (r $\approx$ -0.02-0.11), these superficial metrics are easily manipulated through post-processing and lack the robustness of distribution-based approaches, as discussed in Section~\ref{sec:Background}. A comprehensive analysis of LD-score and 196 other stylometric methods is summarized in Appendix~\ref{sec:stylo}, establishing LD-score as a valuable metric.

\section{Dataset Construction}
\label{sec:dataset}

\subsection{Motivation and Need}
Existing benchmarks for AI-generated text detection each exhibit significant gaps. HC3~\citep{guo-etal-2023-hc3}, while pioneering, covers only ChatGPT-3.5 and lacks multi-model and multi-temperature coverage. M4~\citep{m4} spans multiple domains and models but uses inconsistent model sets across domains and omits temperature variation. Ghostbuster~\citep{ghost} offers prompt-aligned multi-model responses but is limited to $\sim$1,000 samples per domain at a single temperature. RealDet~\citep{realdet} improves scale and breadth but similarly lacks consistent cross-domain model coverage, temperature variation, and relies on relatively weak adversarial attacks. We discuss existing datasets in-depth in Appendix~\ref{app:mdt_analysis}.

These datasets suffer from one or more of the following limitations: (1) reliance on outdated models, (2) lack of prompt-level alignment across generators, (3) absence of temperature variation, (4) limited domain coverage, (5) insufficient scale for robust statistical analysis, or (6) lack of targeted adversarial augmentation beyond standard paraphrasing. We address all of these concerns with our comprehensive dataset construction, which comprises \textbf{642,274} samples spanning five domains, four models, three temperature settings, and three adversarial attack strategies.

\subsection{Dataset Generation}

\textbf{Base Dataset and Human Text.} We used the HC3 corpus \citep{guo-etal-2023-hc3} as our foundation, leveraging its 24,322 unique prompts spanning five domains with 58,546 authentic human responses. This provides crucial domain diversity, ranging from highly technical fields (finance, medicine) to conversational contexts (Reddit\_ELI5), with average response lengths varying from 186.8 to 1,301.6 tokens.

\textbf{AI Model Response Generation.} We generated synthetic responses using four recent mid-sized state-of-the-art open-source models: \textbf{Llama 3.1 8B}, \textbf{Gemma 3 12B}, \textbf{Qwen2.5-VL 7B}, and \textbf{Ministral 8B}. These models represent diverse architectural approaches and training paradigms while remaining computationally accessible for reproduction.

For each of the 24,322 prompts, we generated three responses per model at temperature settings of \textbf{0.2} (deterministic), \textbf{0.5} (balanced), and \textbf{0.8} (stochastic). This temperature stratification is essential, as sampling temperature directly modulates vocabulary probability distributions. An overview of the dataset composition is tabulated in Table~\ref{tab:mdt-domain-composition} in Appendix~\ref{app:mdt_analysis}.

\subsection{Adversarial Augmentation}

To evaluate robustness against evasion, we augmented the dataset with adversarial variants generated by the originating model itself: (A) a standard paraphrase, (B) a paraphrase avoiding a randomly selected letter $\ell_{1}$, and (C) a paraphrase avoiding two distinct letters $\ell_{1} \neq \ell_{2}$. These constraints directly stress-test letter-distribution-based detection by forcing shifts in character-level statistics, and effectively double the dataset size. Attack success analysis is presented in Appendix~\ref{app:mdt_analysis}.



\section{Implementation for LLM Text Detection}
\label{sec:Methodology}

Having discussed the empirical results proving the ``Wall of Separation" between letter distributions of Human and LLM-generated text, we now develop an approach utilizing this finding and set up experiments to analyze improvement in performance of existing training-free black-box methods (\textbf{DNA}, \textbf{Bino} and \textbf{FastDetectGPT} \citep{bao2024fastdetectgptefficientzeroshotdetection}) through an augmentation strategy. 

We consider responses from 3 candidate models to adequately quantify the difference in letter distributions between input text and that of reference AI text. Our scoring mechanism involves combining the text outputs from all 3 candidate models and computing the LD-score (Subsection~\ref{lds}) between the input text and this combined reference distribution.


To retain the full discriminative power of both signals, we instead represent each sample $s$ as a two-dimensional feature vector:
\begin{align}
\mathbf{x}(s) &= \begin{bmatrix} R(s) \\ \mathrm{\text{LD-Score}}\!\left(P_{\text{test}}(s) \parallel P_{\mathcal{M}}(s)\right) \end{bmatrix}
\end{align}
where $R(s)$ denotes the base detector score for sample $s$, $P_{\text{test}}$ is the letter distribution of the input text, and $P_{\mathcal{M}}$ is the letter distribution computed over the combined text outputs of all candidate models in $\mathcal{M}$.

 We employ a Support Vector Machine (SVM) with a radial basis function (RBF) kernel over our 2-dimensional input vector $x(s)$(discussed further in Appendix~\ref{sec:more_method}).

\section{Experiments}
\label{sec:Experiments}


\subsection{Experimental Setup}
\textbf{Datasets.} We set up our experiments by sourcing our data from the MDTA and Ghostbuster datasets, drawing AI samples from 4 different models. For Ghostbuster, we use the following models: ``GPT4 Turbo 2024-04-09", ``GPT4 Omni", ``Claude 3 Opus", and ``GigaChat Pro". We then separate each dataset into 4 different class-balanced sections, one for each of the 4 models serving as the "AI model". The remaining model responses form the three models we draw LD-score from.

\textbf{Baselines} DNA, Bino and FastDetectGPT (FDGPT) are adopted as baseline training-free methods and are augmented with the LD-Score. “Falcon-7b-Instruct” and “Falcon-7b” serve as the reference (performer) and observer models, respectively, in DNA and Bino approaches. FDGPT uses GPT-J-6B for sampling and GPT-Neo-2.7B for scoring.

\subsection{Performance Analysis}

Table~\ref{tab:main_results} reports F1 and AUROC for baselines and their LD-Score 
augmentations with only 100 training samples (50 human, 50 AI). LD-DNA achieves the 
highest average F1 (0.94) and AUROC (0.97), with consistent improvements over DNA 
across nearly all domains. The gains are largest in structured domains—Reuters 
($\Delta$F1: +0.02), Reddit ELI5 ($\Delta$F1: +0.04), and Open QA ($\Delta$F1: 
+0.03)—where domain-specialized vocabularies amplify the LD signal, consistent with 
the theoretical framework in Section~\ref{sec:The Letter Distribution Score}. Finance and Wiki CSAI show 
minimal gains, as their broader vocabularies weaken distributional separation. Notably, 
LD-Score augmentation consistently outperforms perplexity-only ensembles (DNA+Bino, 
DNA+FastDetectGPT), demonstrating that the orthogonality of LD-Score to perplexity 
signals provides complementary discriminative power beyond what combining perplexity 
methods alone achieves. Across all methods, augmentation with LD-Score also reduces 
variance, indicating more stable detection under limited training data.

\begin{table*}[t]
\centering
\resizebox{\textwidth}{!}{%
\begin{tabular}{l | c | ccccc | c}
\toprule
& \textbf{Ghostbuster} & \multicolumn{5}{c|}{\textbf{MDTA}} & \\
\textbf{Method} & Avg & Finance & Medicine & Open QA & Reddit ELI5 & Wiki CSAI & Avg \\
\midrule
\multicolumn{8}{l}{\textit{AUROC}} \\
\midrule
DNA & 0.994{\tiny{$\pm$0.005}} & 0.958{\tiny{$\pm$0.016}} & 0.996{\tiny{$\pm$0.005}} & 0.812{\tiny{$\pm$0.043}} & 0.971{\tiny{$\pm$0.008}} & \textbf{0.992}\tiny{$\pm$0.006} & 0.946{\tiny{$\pm$0.021}} \\
Bino & 0.985{\tiny{$\pm$0.008}} & 0.946{\tiny{$\pm$0.014}} & 0.995{\tiny{$\pm$0.006}} & 0.817{\tiny{$\pm$0.042}} & 0.966{\tiny{$\pm$0.009}} & 0.990{\tiny{$\pm$0.008}} & 0.943{\tiny{$\pm$0.021}} \\
FDGPT & 0.935{\tiny{$\pm$0.019}} & 0.871{\tiny{$\pm$0.011}} & 0.954{\tiny{$\pm$0.011}} & 0.774{\tiny{$\pm$0.048}} & 0.930{\tiny{$\pm$0.011}} & 0.963{\tiny{$\pm$0.011}} & 0.898{\tiny{$\pm$0.024}} \\
\midrule
LD-DNA & \textbf{0.995}{\tiny{$\pm$0.005}} & \textbf{0.974}\tiny{$\pm$0.008} & \textbf{0.998}\tiny{$\pm$0.001} & 0.856{\tiny{$\pm$0.029}} & \textbf{0.982}\tiny{$\pm$0.007} & \textbf{0.992}\tiny{$\pm$0.004} & \textbf{0.960}\tiny{$\pm$0.014} \\
LD-Bino & 0.989{\tiny{$\pm$0.005}} & 0.970{\tiny{$\pm$0.006}} & 0.997{\tiny{$\pm$0.003}} & \textbf{0.859}\tiny{$\pm$0.031} & 0.978{\tiny{$\pm$0.008}} & 0.990{\tiny{$\pm$0.005}} & 0.959{\tiny{$\pm$0.015}} \\
LD-FDGPT & 0.960{\tiny{$\pm$0.016}} & 0.888{\tiny{$\pm$0.014}} & 0.961{\tiny{$\pm$0.010}} & 0.808{\tiny{$\pm$0.026}} & 0.934{\tiny{$\pm$0.015}} & 0.956{\tiny{$\pm$0.011}} & 0.909{\tiny{$\pm$0.016}} \\
\midrule
DNA+Bino & 0.993{\tiny{$\pm$0.005}} & \textbf{0.974}\tiny{$\pm$0.008} & 0.994{\tiny{$\pm$0.004}} & 0.832{\tiny{$\pm$0.039}} & 0.958{\tiny{$\pm$0.009}} & 0.989{\tiny{$\pm$0.011}} & 0.949{\tiny{$\pm$0.019}} \\
DNA+FDGPT & \textbf{0.995}\tiny{$\pm$0.004} & 0.971{\tiny{$\pm$0.010}} & 0.996{\tiny{$\pm$0.003}} & 0.826{\tiny{$\pm$0.038}} & 0.965{\tiny{$\pm$0.010}} & 0.990{\tiny{$\pm$0.006}} & 0.950{\tiny{$\pm$0.018}} \\
\midrule
\multicolumn{8}{l}{\textit{F1 Score}} \\
\midrule
DNA & 0.971{\tiny{$\pm$0.014}} & \textbf{0.933}\tiny{$\pm$0.011} & 0.976{\tiny{$\pm$0.009}} & 0.756{\tiny{$\pm$0.040}} & 0.913{\tiny{$\pm$0.026}} & \textbf{0.958}\tiny{$\pm$0.014} & 0.907{\tiny{$\pm$0.023}} \\
Bino & 0.943{\tiny{$\pm$0.027}} & 0.905{\tiny{$\pm$0.019}} & 0.977{\tiny{$\pm$0.009}} & 0.752{\tiny{$\pm$0.048}} & 0.895{\tiny{$\pm$0.024}} & 0.949{\tiny{$\pm$0.017}} & 0.896{\tiny{$\pm$0.027}} \\
FDGPT & 0.864{\tiny{$\pm$0.032}} & 0.751{\tiny{$\pm$0.043}} & 0.879{\tiny{$\pm$0.028}} & 0.701{\tiny{$\pm$0.042}} & 0.832{\tiny{$\pm$0.035}} & 0.896{\tiny{$\pm$0.025}} & 0.812{\tiny{$\pm$0.035}} \\
\midrule
LD-DNA & \textbf{0.980}\tiny{$\pm$0.007} & 0.929{\tiny{$\pm$0.017}} & \textbf{0.982}\tiny{$\pm$0.006} & \textbf{0.788}\tiny{$\pm$0.027} & \textbf{0.949}\tiny{$\pm$0.012} & 0.955{\tiny{$\pm$0.010}} & \textbf{0.921}\tiny{$\pm$0.016} \\
LD-Bino & 0.965{\tiny{$\pm$0.013}} & 0.918{\tiny{$\pm$0.017}} & 0.976{\tiny{$\pm$0.008}} & 0.784{\tiny{$\pm$0.028}} & 0.936{\tiny{$\pm$0.013}} & 0.956{\tiny{$\pm$0.008}} & 0.914{\tiny{$\pm$0.017}} \\
LD-FDGPT & 0.916{\tiny{$\pm$0.022}} & 0.770{\tiny{$\pm$0.032}} & 0.907{\tiny{$\pm$0.013}} & 0.720{\tiny{$\pm$0.027}} & 0.856{\tiny{$\pm$0.015}} & 0.885{\tiny{$\pm$0.018}} & 0.828{\tiny{$\pm$0.022}} \\
\midrule
DNA+Bino & 0.971{\tiny{$\pm$0.008}} & 0.927{\tiny{$\pm$0.015}} & 0.979{\tiny{$\pm$0.005}} & 0.764{\tiny{$\pm$0.044}} & 0.911{\tiny{$\pm$0.018}} & \textbf{0.958}\tiny{$\pm$0.011} & 0.908{\tiny{$\pm$0.023}} \\
DNA+FDGPT & 0.976{\tiny{$\pm$0.010}} & 0.925{\tiny{$\pm$0.017}} & 0.978{\tiny{$\pm$0.006}} & 0.757{\tiny{$\pm$0.052}} & 0.914{\tiny{$\pm$0.019}} & 0.957{\tiny{$\pm$0.011}} & 0.906{\tiny{$\pm$0.026}} \\
\bottomrule
\end{tabular}%
}
\caption{Detection performance with 100 balanced training samples (50 AI + 50 human) for SVM training and threshold calibration. The Ghostbuster column reports the average AUROC/F1 across Essay, Reuters, and WP domains. LD-X denotes augmentation of method X with the LD-Score via RBF-SVM fusion. Results are mean $\pm$ std over 5 runs with different seeds. \textbf{Bold} indicates best per column.}
\label{tab:main_results}
\end{table*}

\textbf{Unbalanced Training Regime} To further stress-test LD-Score fusion, we evaluate under a more challenging unbalanced training regime where human samples are scarce (Table~\ref{tab:sample_efficiency}).
In this setting, DNA and Binoculars struggle to learn a reliable decision boundary, improving only gradually with more data (DNA: $0.847 \to 0.886$). By contrast, their LD-Score-augmented counterparts converge rapidly from the outset, achieving AUROC of 0.935 and 0.918 with only 100 samples, gains of $+0.088$ and $+0.122$ respectively. Augmented methods plateau early and maintain their advantage throughout, with the diminishing $\Delta$ at larger sample sizes reflecting baselines slowly catching up via improved threshold calibration rather than any degradation of the fusion. This is particularly significant for real-world AI detection, where human-generated text is harder to collect and label at scale. Furthermore, these results reveal that the LD-Score requires surprisingly few domain examples to capture a domain's letter distribution, demonstrating that domain specialization does not pose a practical obstacle to deployment.

\begin{table}[t]
\centering
\begin{tabular}{ccccccc}
\toprule
\textbf{Train} & \textbf{DNA} & \textbf{LD-DNA} & $\boldsymbol{\Delta}$ & \textbf{Bino} & \textbf{LD-Bino} & $\boldsymbol{\Delta}$ \\
\midrule
100  & 0.847 & 0.935 & \textbf{+0.088} & 0.796 & 0.918 & \textbf{+0.122} \\
250  & 0.863 & 0.949 & \textbf{+0.086} & 0.823 & 0.945 & \textbf{+0.122} \\
500  & 0.864 & 0.945 & \textbf{+0.081} & 0.812 & 0.942 & \textbf{+0.130} \\
1000 & 0.886 & 0.945 & \textbf{+0.059} & 0.801 & 0.935 & \textbf{+0.134} \\
\bottomrule
\end{tabular}
\caption{AUROC under an unbalanced training regime (human samples scarce) as a
function of training set size. DNA and Binoculars improve only gradually with more
data, while their LD-Score-augmented counterparts (LD-DNA, LD-Bino) converge rapidly and
maintain a consistent advantage throughout.}
\label{tab:sample_efficiency}
\end{table}

\textbf{{Adversarial Experiments}.} The adversarial variants tested here represent a stress test of our augmentation pipeline, since the LD-Score is derived from clean, non-adversarial model responses and is never optimized against these attacks; full results are shown in the Appendix~\ref{app:adv}. Despite this, LD-Score augmentation improves AUROC over the base detectors in nearly every condition, matching or exceeding base performance in 29 of 30 domain-attack comparisons. Gains are consistent across all three attack types, with average AUROC improvements of $+0.005$ and $+0.010$ for DNA and Binoculars respectively under paraphrase attacks (A), $+0.006$ and $+0.014$ under single letter removal (B), and $+0.006$ and $+0.012$ under dual letter removal (C). Although F1 score improvements are less consistent than AUROC gains, the augmented variants remain better than or broadly comparable to their unaugmented counterparts.

\section{Conclusion and Limitations}

This work introduces letter distribution signatures as an orthogonal detection signal for AI-generated text, establishing a ``Wall of Separation'' where human-AI divergence systematically exceeds AI-AI divergence. Integration with state-of-the-art methods yields consistent AUROC and F1 improvements, with LD-DNA achieving average F1 of 0.921 and AUROC of 0.960 compared to 0.907 and 0.946 for DNA alone, and low correlation (r=0.08-0.13) with perplexity-based methods confirming orthogonality. Key limitations include domain dependence, where the signal is strongest in specialized domains and weaker in open-domain settings, and reliance on surrogate LLMs, which introduces computational overhead. Future work should explore stronger fusion strategies, more comprehensive adversarial robustness including persona prompting and sophisticated paraphrasing attacks, training data contamination detection via distribution matching, and extensions to AI image detection through spectral distribution analysis.

\bibliography{References/reference}
\bibliographystyle{colm2026_conference}

\newpage
\appendix




%
\clearpage
\section{Methodology Details}

\subsection{Algorithm}
Algorithm~\ref{alg:ld_score} details the computation of the Letter Distribution Score between two text samples. Given texts $T_1$ and $T_2$, we first extract normalized letter frequency probability distributions $P_{T_1}$ and $P_{T_2}$ over the 26-character English alphabet. We then compute the Jensen-Shannon Divergence (JSD) between these distributions, a symmetric and smoothed variant of the Kullback-Leibler divergence that avoids the numerical instability arising when one distribution assigns zero probability to a letter. Taking the square root yields the Root Jensen-Shannon Distance (RJSD), which satisfies the triangle inequality and thus constitutes a proper metric space - a desirable formal property that raw JSD does not provide. The resulting LD-Score is bounded in $[0, 1]$, where values near zero indicate near-identical letter distributions and values approaching one indicate maximally divergent distributions.

\begin{algorithm}[t]
\caption{Letter Distribution Score Computation}
\label{alg:ld_score}
\begin{algorithmic}[1]
\INPUT Two text samples $T_1$ and $T_2$
\OUTPUT LD-Score between $T_1$ and $T_2$
\STATE
\STATE // Extract letter distributions
\FOR{each text $t \in \{T_1, T_2\}$}
    \FOR{each letter $\ell \in \{a, \ldots, z\}$}
        \STATE $P_t(\ell) \leftarrow \frac{\text{count}(\ell \text{ in } t)}{\sum_{\ell'} \text{count}(\ell' \text{ in } t)}$
    \ENDFOR
\ENDFOR
\STATE
\STATE // Compute Jensen-Shannon Divergence
\STATE $M \leftarrow \frac{1}{2}(P_{T_1} + P_{T_2})$
\STATE $\text{KL}(P \| Q) \leftarrow \sum_{i=1}^{26} P(i) \log \frac{P(i)}{Q(i)}$ \COMMENT{Kullback-Leibler divergence~\citep{kullback1951}}
\STATE $\text{JSD}(P_{T_1} \| P_{T_2}) \leftarrow \frac{1}{2}[\text{KL}(P_{T_1} \| M) + \text{KL}(P_{T_2} \| M)]$ \COMMENT{Jensen-Shannon divergence~\citep{lin1991}}
\STATE
\STATE // Compute Root Jensen-Shannon Distance
\textbf{return} $\text{LD-Score}(T_1, T_2) = \sqrt{\text{JSD}(P_{T_1} \| P_{T_2})}$ \COMMENT{RJSD metric~\cite{endres2003}}
\end{algorithmic}
\end{algorithm}

\subsection{SVM Implementation}
\label{sec:more_method}
The SVM operates using a radial basis function (RBF) kernel to capture the non-linear decision boundary:
\begin{align}
\hat{y}(s) &= \mathrm{sign}\!\left(\sum_{i} \alpha_i y_i \, K(\mathbf{x}_i, \mathbf{x}(s)) + b\right)
\end{align}
where $K(\mathbf{x}_i, \mathbf{x}(s)) = \exp\!\left(-\gamma \|\mathbf{x}_i - \mathbf{x}(s)\|^2\right)$ is the RBF kernel.

This formulation allows the classifier to learn curved, non-linear boundaries in the two-dimensional feature space, effectively exploiting the complementary structure of both detection signals. While we use DNA and Binoculars as baselines here, this approach generalizes to any perplexity-based method paired with the LD-score.

\clearpage

\section{Ghostbuster Analysis}
\label{app:domain_analysis}

\subsection{Domain-Specific Dataset Analysis}

This appendix provides detailed pairwise Jensen-Shannon distance matrices and hierarchical clustering dendrograms for individual domains, complementing the essay domain analysis presented in the main text (Section~\ref{sec:empirical_validation}).

\begin{figure}[h]
    \centering
    \begin{subfigure}[b]{0.32\textwidth}
        \centering
        \includegraphics[width=\textwidth]{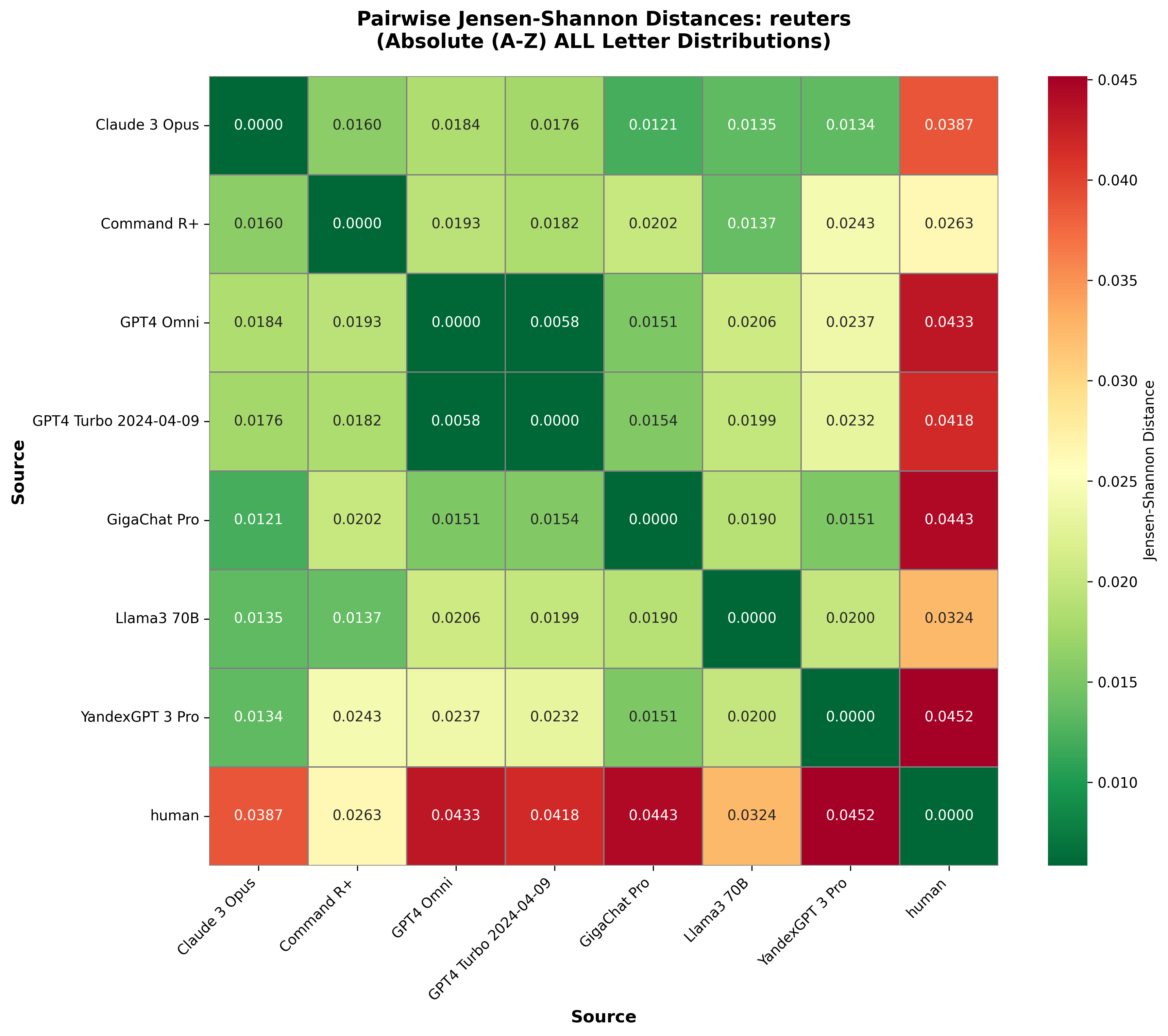}
        \caption{Reuters}
        \label{fig:reuters_pairwise}
    \end{subfigure}
    \hfill
    \begin{subfigure}[b]{0.32\textwidth}
        \centering
        \includegraphics[width=\textwidth]{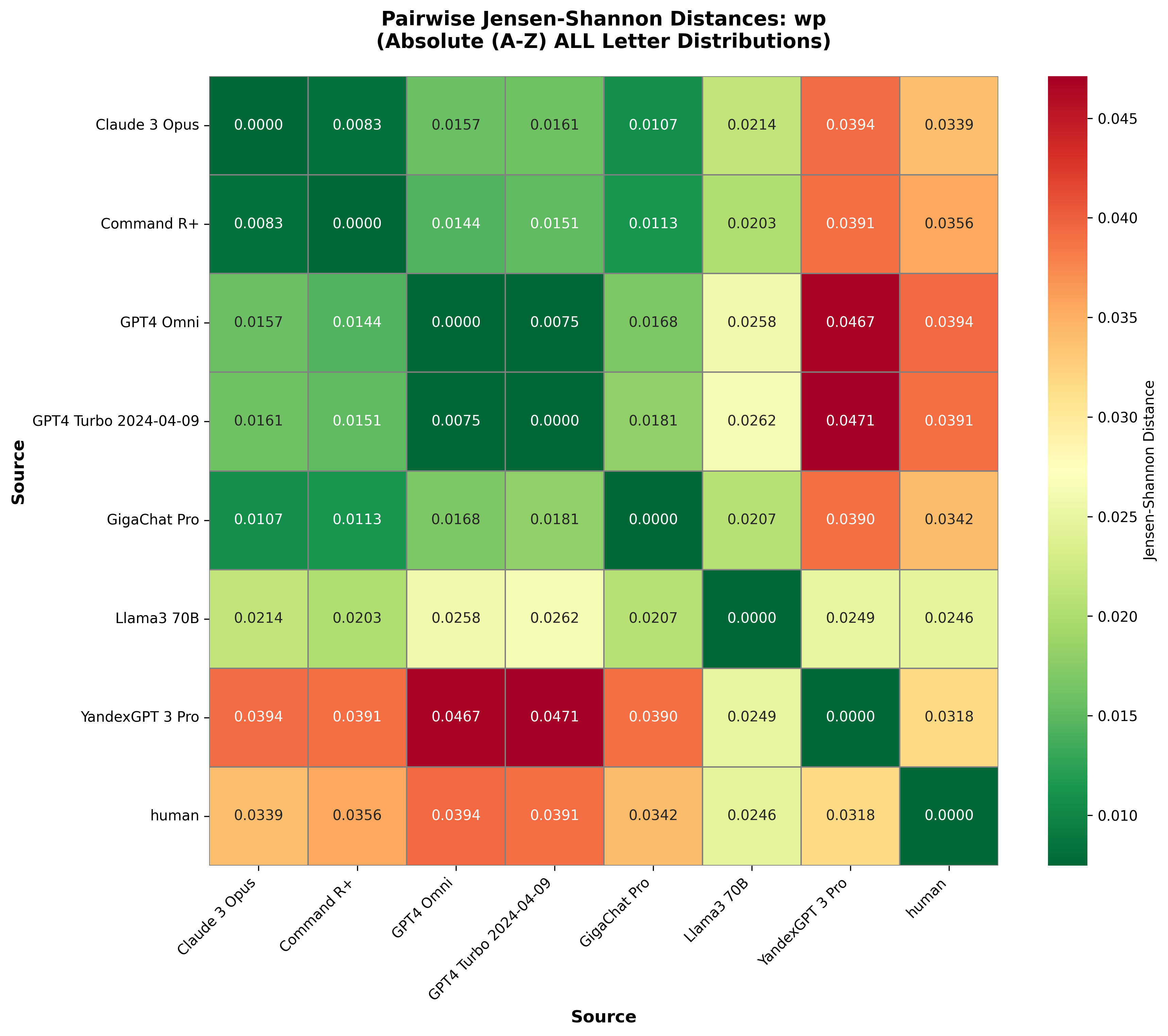}
        \caption{Creative Writing}
        \label{fig:creative_pairwise}
    \end{subfigure}
    \hfill
    \begin{subfigure}[b]{0.32\textwidth}
        \centering
        \includegraphics[width=\textwidth]{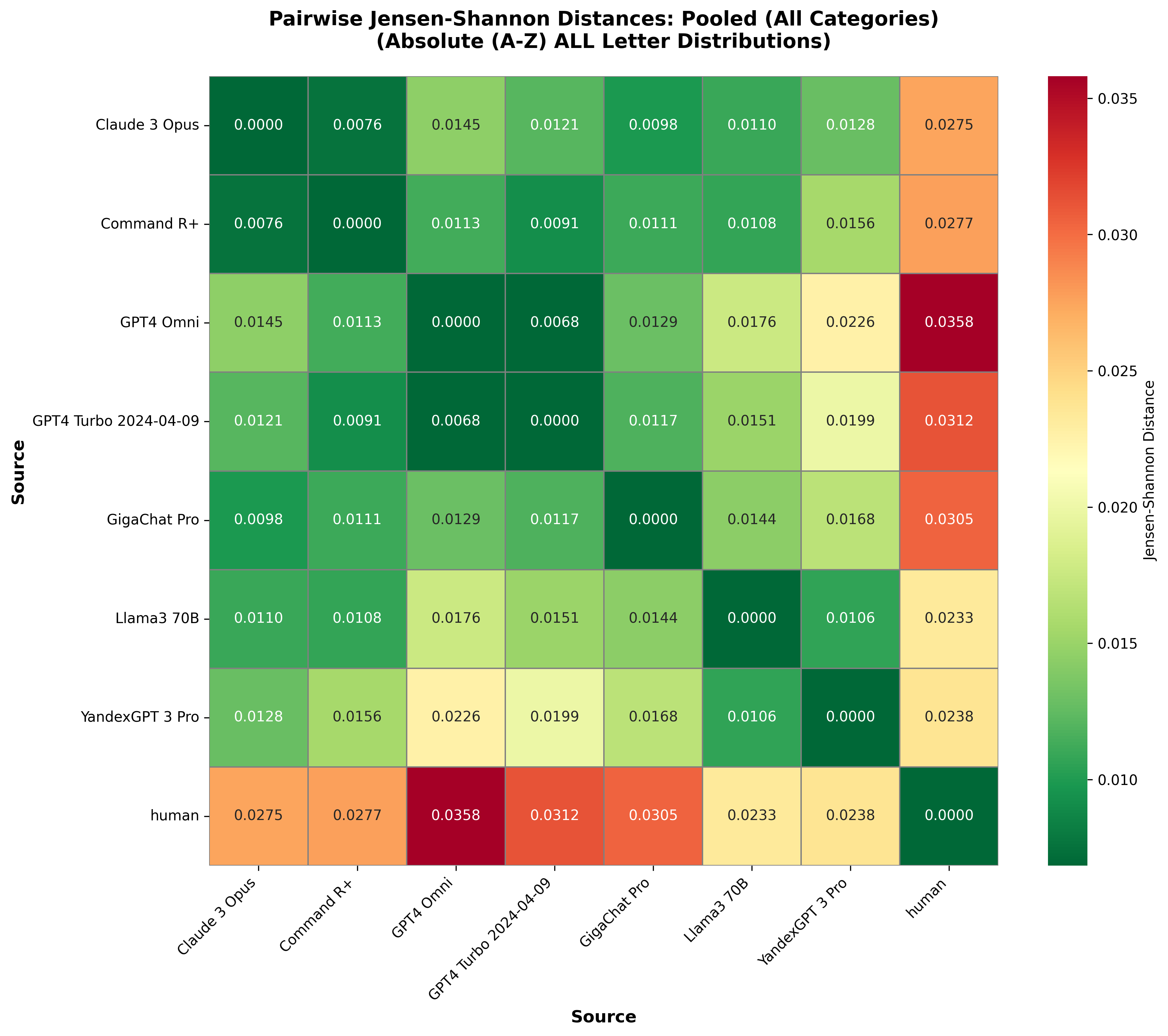}
        \caption{Pooled}
        \label{fig:pooled_pairwise}
    \end{subfigure}
    
    \vspace{0.5cm}
    
    \begin{subfigure}[b]{0.32\textwidth}
        \centering
        \includegraphics[width=\textwidth]{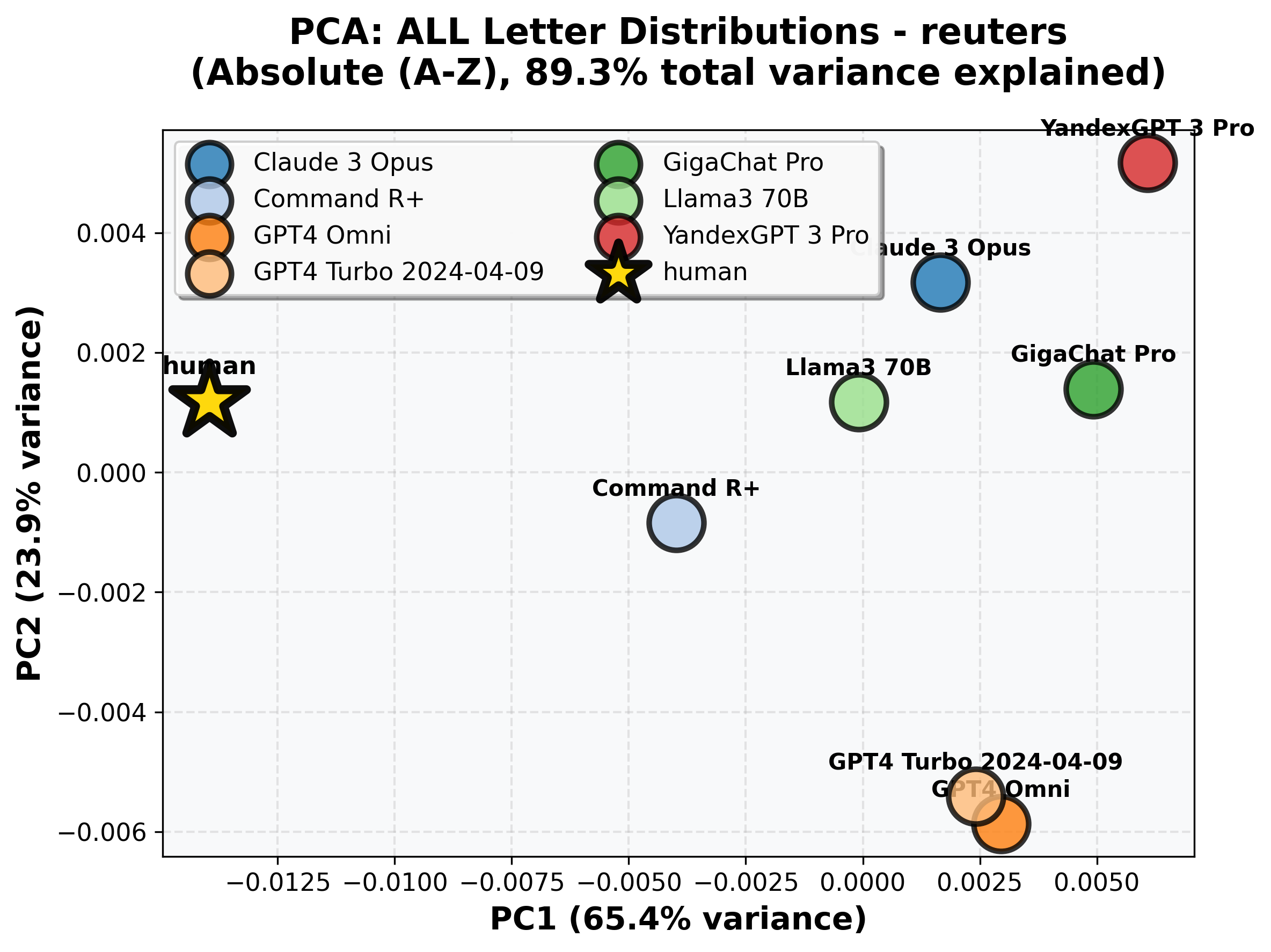}
        \caption{Reuters}
        \label{fig:reuters_pca}
    \end{subfigure}
    \hfill
    \begin{subfigure}[b]{0.32\textwidth}
        \centering
        \includegraphics[width=\textwidth]{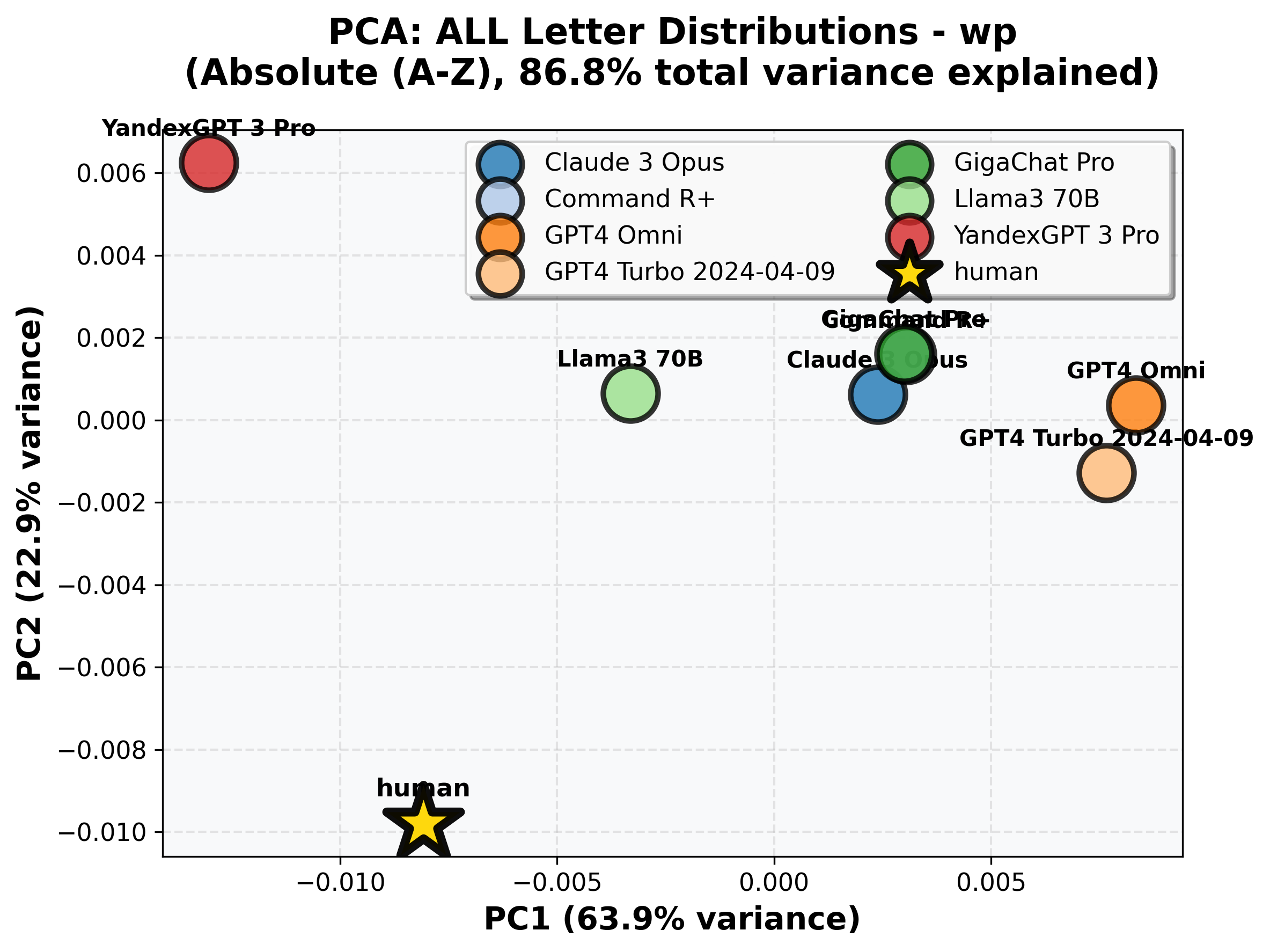}
        \caption{Creative Writing}
        \label{fig:creative_pca}
    \end{subfigure}
    \hfill
    \begin{subfigure}[b]{0.32\textwidth}
        \centering
        \includegraphics[width=\textwidth]{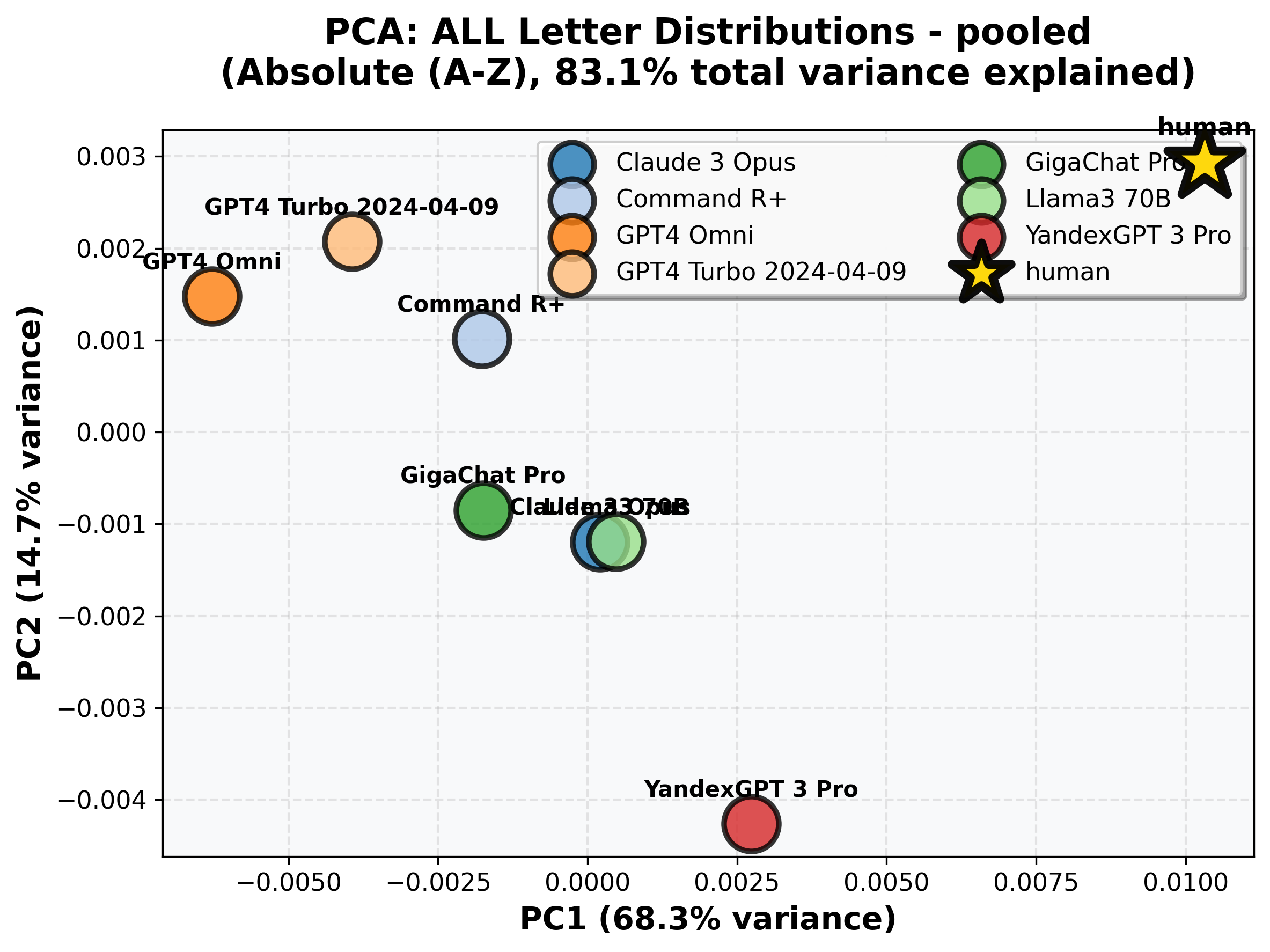}
        \caption{Pooled}
        \label{fig:pooled_pca}
    \end{subfigure}
    
    \vspace{0.5cm}
    
    \begin{subfigure}[b]{0.32\textwidth}
        \centering
        \includegraphics[width=\textwidth]{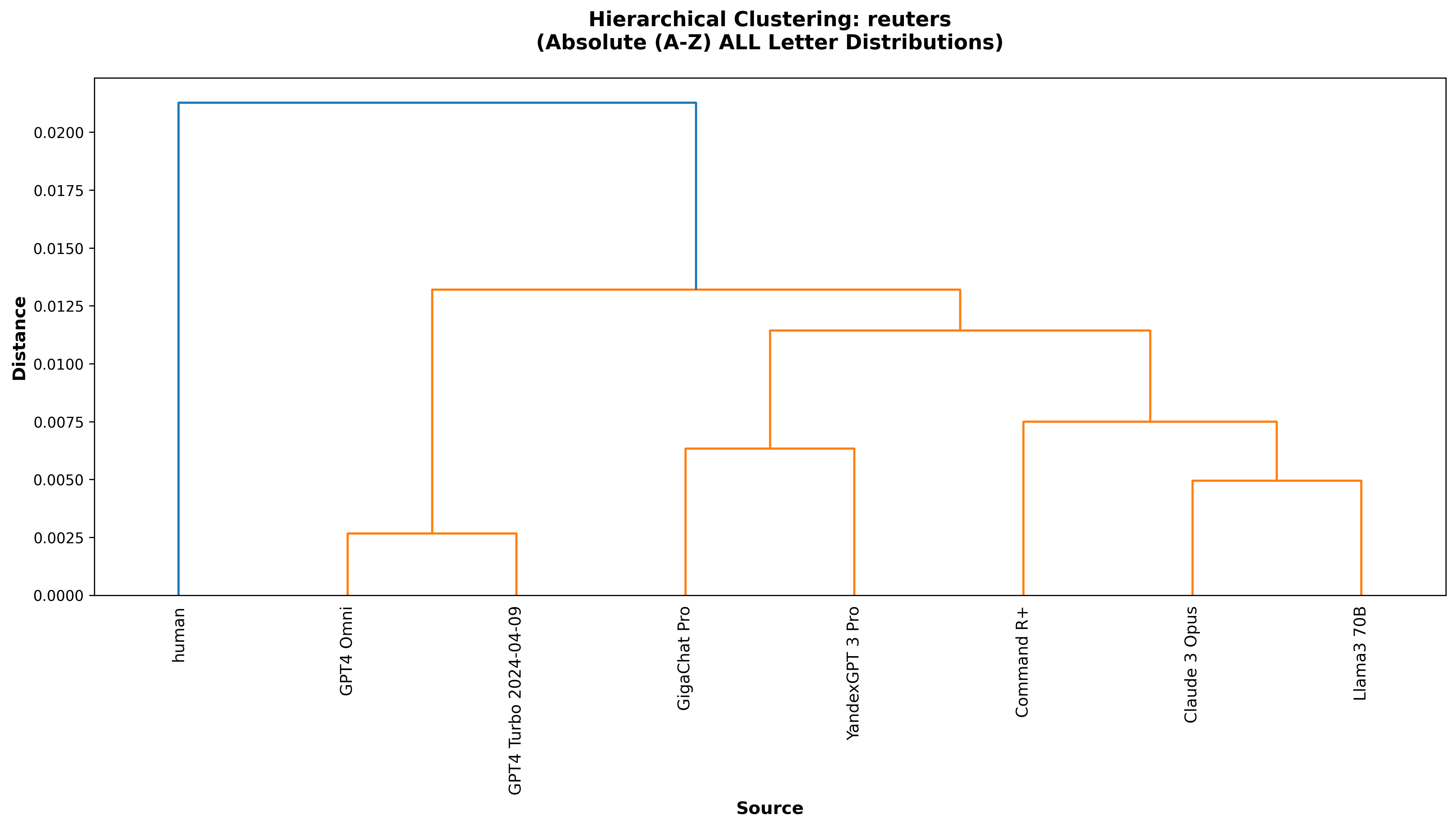}
        \caption{Reuters}
        \label{fig:reuters_dendrogram}
    \end{subfigure}
    \hfill
    \begin{subfigure}[b]{0.32\textwidth}
        \centering
        \includegraphics[width=\textwidth]{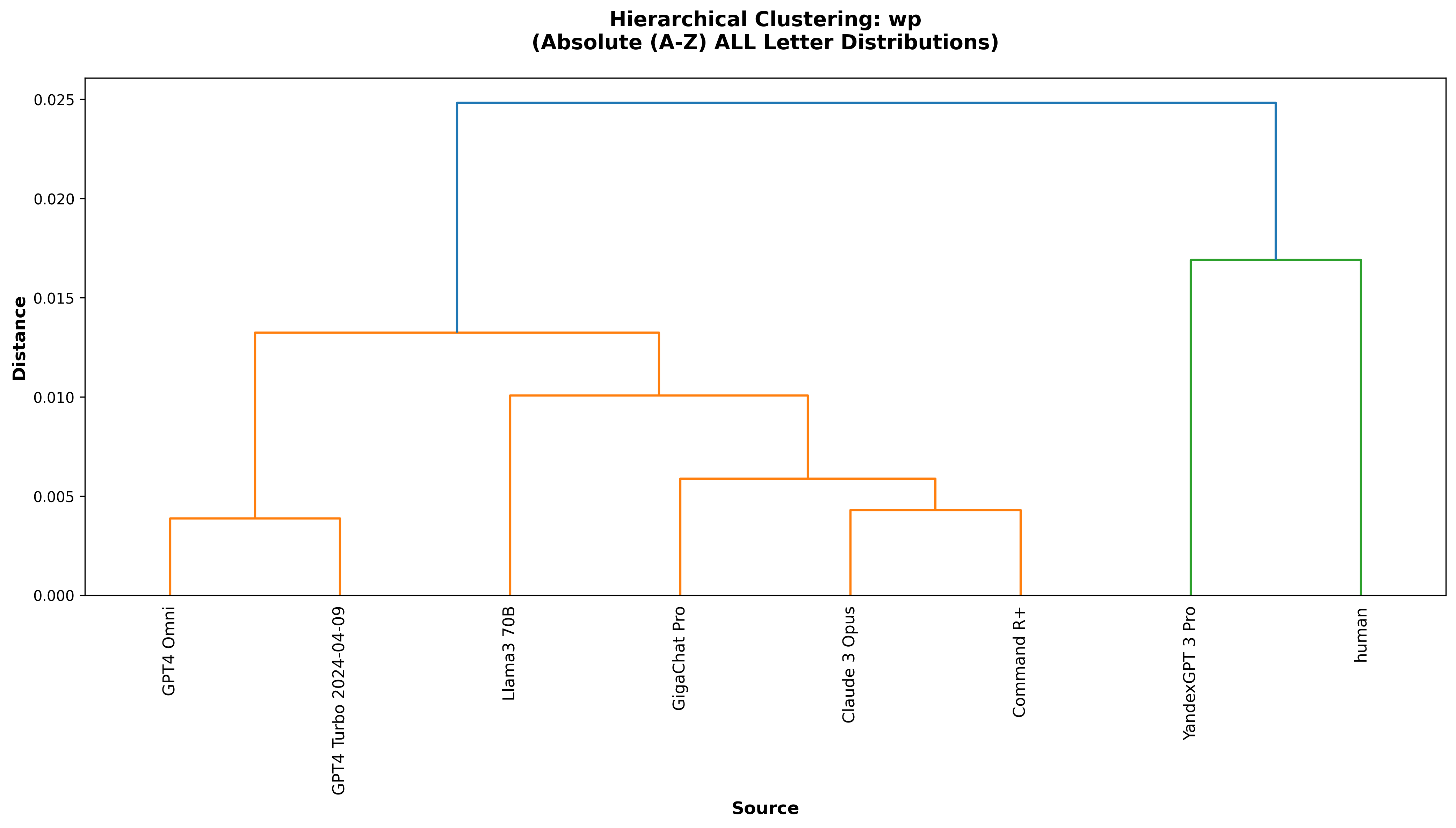}
        \caption{Creative Writing}
        \label{fig:creative_dendrogram}
    \end{subfigure}
    \hfill
    \begin{subfigure}[b]{0.32\textwidth}
        \centering
        \includegraphics[width=\textwidth]{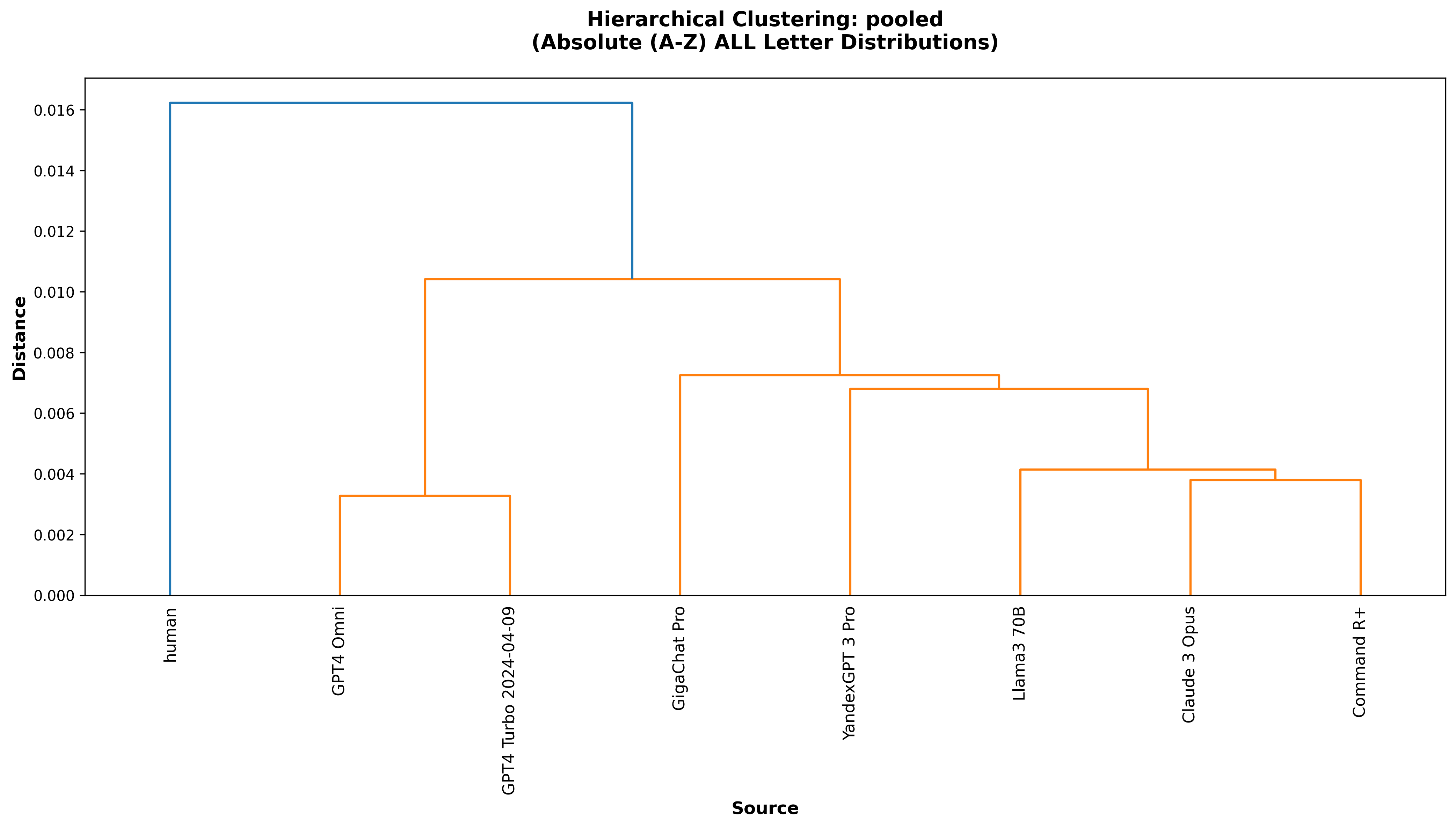}
        \caption{Pooled}
        \label{fig:pooled_dendrogram}
    \end{subfigure}
    
    \caption{Domain-specific LD-Score analysis on Ghostbuster dataset. Top row: Pairwise Jensen-Shannon distance matrices. Middle row: PCA projections visualizing geometric separation. Bottom row: Hierarchical clustering dendrograms. Reuters (specialized news) shows strongest separation, Creative Writing (general, unstructured) shows weakest separation, and Pooled results demonstrate overall robustness.}
    \label{fig:ghostbuster_domain_analysis}
\end{figure}

The hierarchical clustering dendrograms (bottom row of Figure~\ref{fig:ghostbuster_domain_analysis}) reveal the predicted clustering structure. Across all domains, similar models cluster together, with GPT-4 family models showing particularly tight grouping. Critically, human text mostly splits at the top level of the hierarchy, forming a completely separate cluster from all AI models. This top-level split validates our theoretical prediction that human text occupies a fundamentally distinct region in letter distribution space, separated by the Wall of Separation from the AI model cluster.

\subsection{Domain-Dependent Tightness of the AI Cluster}
\label{sec:domain_cluster_tightness}

\begin{figure}[h]
    \centering
    \includegraphics[width=0.75\linewidth]{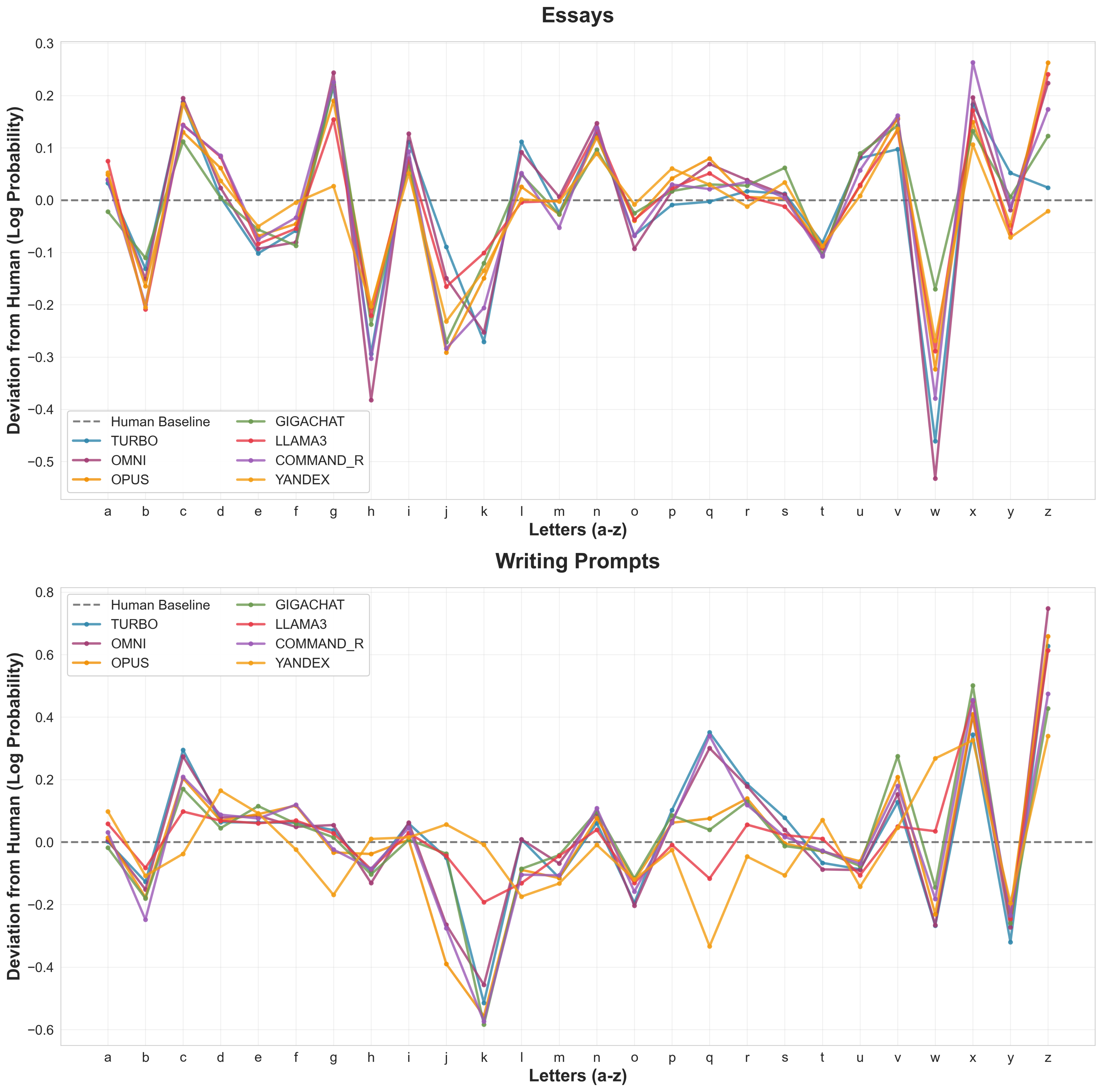}
    \caption{Letter-level log-probability deviations from the human baseline across two domains: Essays (top) and Writing Prompts (bottom). Curves correspond to different AI models. In structured domains such as essays, AI models exhibit tightly clustered letter distributions, indicating strong AI--AI similarity. In contrast, open-ended creative writing prompts induce greater variability across models, illustrating how domain restrictiveness controls the tightness of the AI cluster and the strength of separation from human text.}
    \label{fig:log_probs}
\end{figure}

Figure~\ref{fig:log_probs} illustrates letter-level log-probability deviations from the human baseline across two domains: Essays and Writing Prompts. In structured domains such as essays, AI-generated texts exhibit highly consistent letter-level deviations across models, resulting in tightly overlapping curves. This reflects strong AI--AI clustering: constrained task structure, formal tone, and standardized vocabulary encourage all models to sample similarly from their shared approximation of the global language distribution.

In contrast, creative domains such as writing prompts induce substantially higher variability across AI models. Open-ended generation amplifies stylistic choices, narrative voice, and lexical experimentation, increasing divergence both among AI models and relative to the human baseline. This expansion of the AI cluster reduces the AI--AI similarity margin and weakens separation, consistent with the domain bias term in Eq. \ref{eq:total_divergence}.

\clearpage
\section{The MDTA Dataset}
\label{app:mdt_analysis}

\subsection{Existing Datasets}
Current datasets for AI-generated text detection suffer from critical limitations that hinder the development of robust, generalizable detection methods. The original HC3 (Human ChatGPT Comparison Corpus) dataset \citep{guo-etal-2023-hc3}, while pioneering in providing domain-diverse human-AI text pairs, contains only ChatGPT-3.5 responses paired with human text. Given the rapid advancement in language model capabilities since early 2023, this dataset is now outdated and fails to capture the linguistic characteristics of modern state-of-the-art models. More fundamentally, it lacks the multi-model and multi-temperature coverage necessary for developing detection methods that generalize across different AI systems and generation strategies.

Beyond HC3, existing benchmark datasets exhibit complementary but insufficient coverage for comprehensive distributional analysis. The M4 benchmark dataset~\citep{m4} is one of the most comprehensive efforts to date, spanning multiple domains and including responses from a wide range of language models. However, it does not consistently employ the same set of models across all domains, which prevents proper cross-domain comparisons of model-specific characteristics. More critically, the models included in M4 are largely outdated, having since been succeeded by substantially more capable language models on both the proprietary and open-source fronts. Additionally, M4 lacks temperature variation in its generation strategy and its adversarial attack configurations, while valuable at the time of release, have become less representative of modern evasion techniques.

The Ghostbuster dataset \citep{ghost} addresses several of these shortcomings, providing prompt-aligned multi-model responses across three domains: creative writing (Writing Prompts), news (Reuters), and student essays. However, it provides only approximately 1,000 samples per domain and generates all responses at a single temperature setting. This limits the dataset's utility for studying how decoding stochasticity influences the statistical properties of generated text more broadly, as low temperatures produce deterministic, repetitive outputs while high temperatures yield greater lexical and structural variation. 

More recently, RealDet~\citep{realdet} introduced a large and comprehensive benchmark spanning many domains, prompts, and LLMs, representing an important step forward in dataset scale and breadth. However, it is less suitable for controlled distributional comparison. The dataset does not consistently provide the same set of LLMs across all domains, which makes systematic cross-domain, cross-model comparisons more difficult. In addition, many of the included models have since been surpassed by substantially stronger proprietary and open-source systems, limiting its usefulness as a benchmark for studying the behavior of current-generation LLMs. RealDet also does not explicitly vary generation temperature, preventing analysis of how decoding stochasticity affects textual distributions. Finally, although it includes adversarial attacks, these largely rely on relatively standard paraphrasing and token-level perturbations, which are weaker than more modern adaptive attacks in which the source LLM itself is instructed to rewrite text under targeted lexical constraints.

\subsection{Domain-Specific Analysis}

\begin{figure}[h]
    \centering
    \begin{subfigure}[b]{0.24\textwidth}
        \centering
        \includegraphics[width=\textwidth]{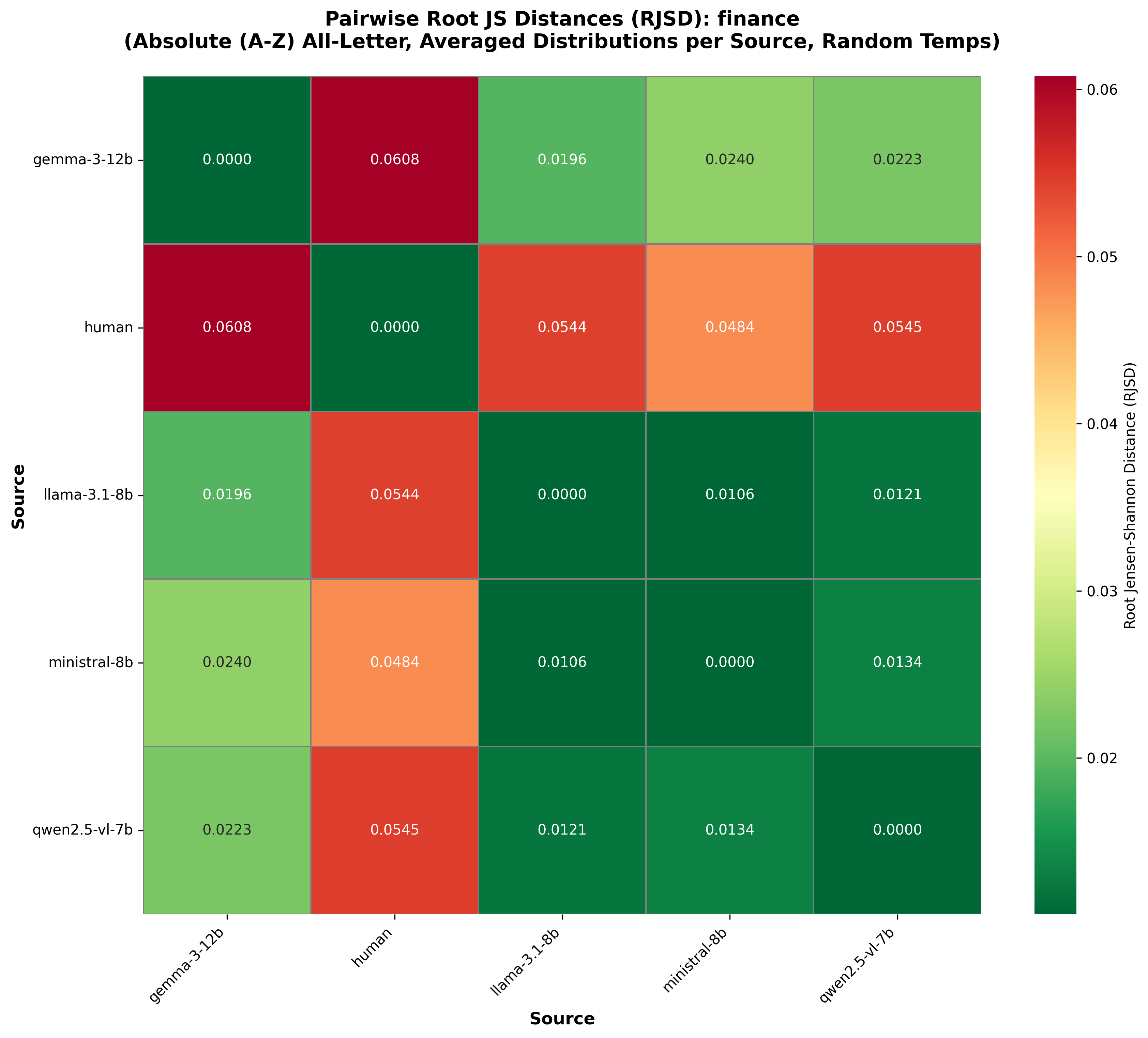}
        \caption{Finance}
        \label{fig:mdt_finance_pairwise}
    \end{subfigure}
    \hfill
    \begin{subfigure}[b]{0.24\textwidth}
        \centering
        \includegraphics[width=\textwidth]{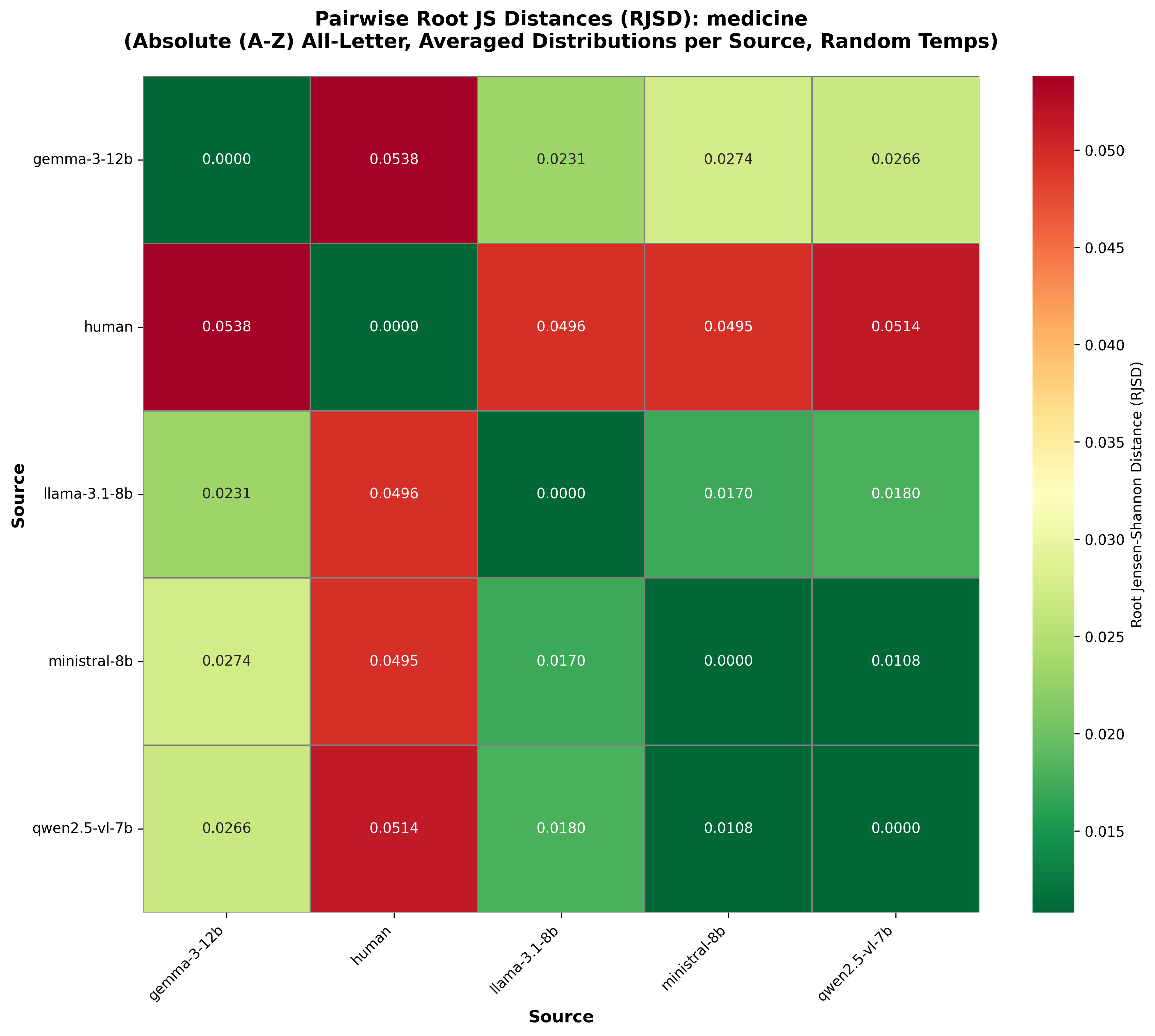}
        \caption{Medicine}
        \label{fig:mdt_medicine_pairwise}
    \end{subfigure}
    \hfill
    \begin{subfigure}[b]{0.24\textwidth}
        \centering
        \includegraphics[width=\textwidth]{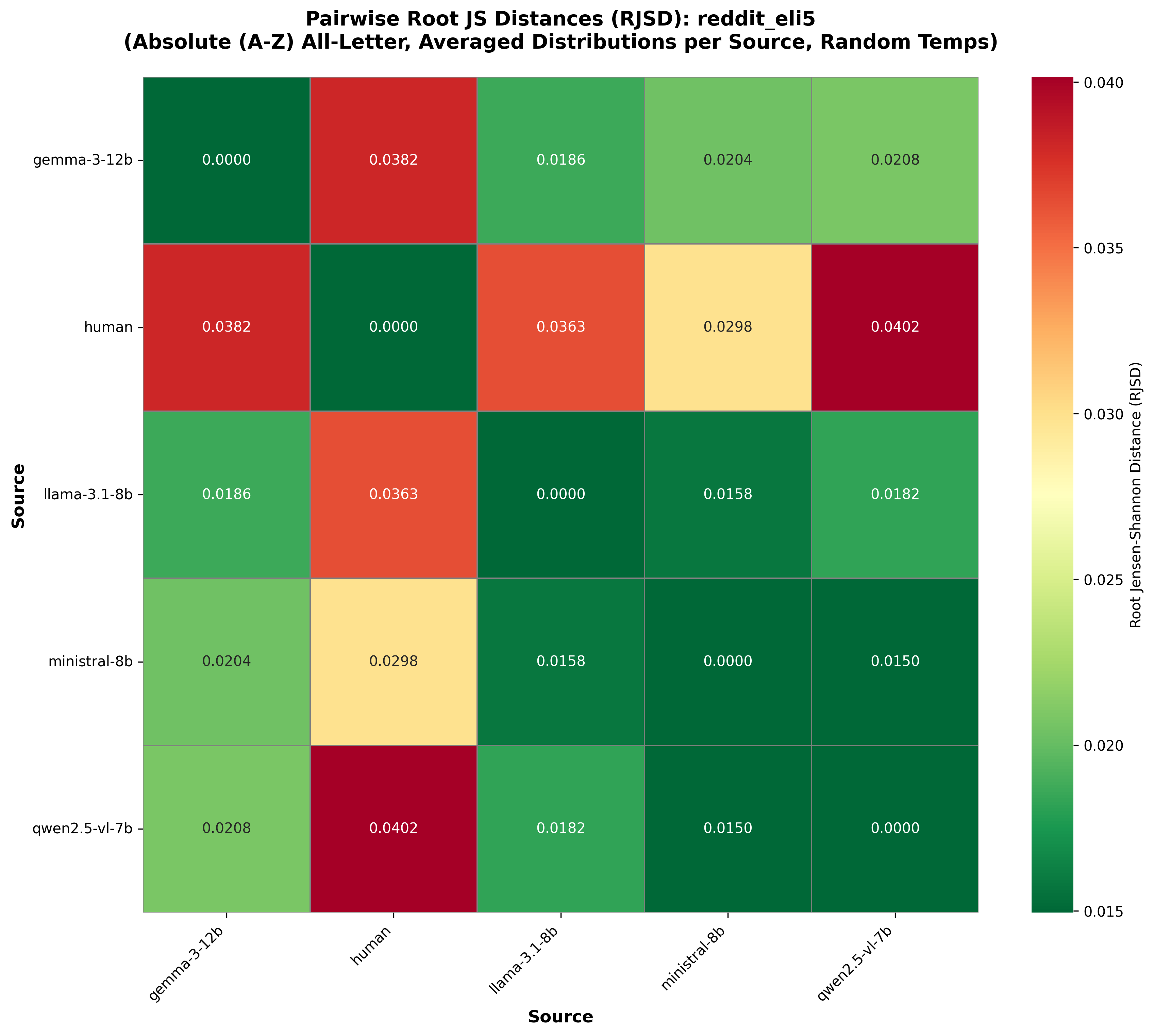}
        \caption{Reddit ELI5}
        \label{fig:mdt_reddit_pairwise}
    \end{subfigure}
    \hfill
    \begin{subfigure}[b]{0.24\textwidth}
        \centering
        \includegraphics[width=\textwidth]{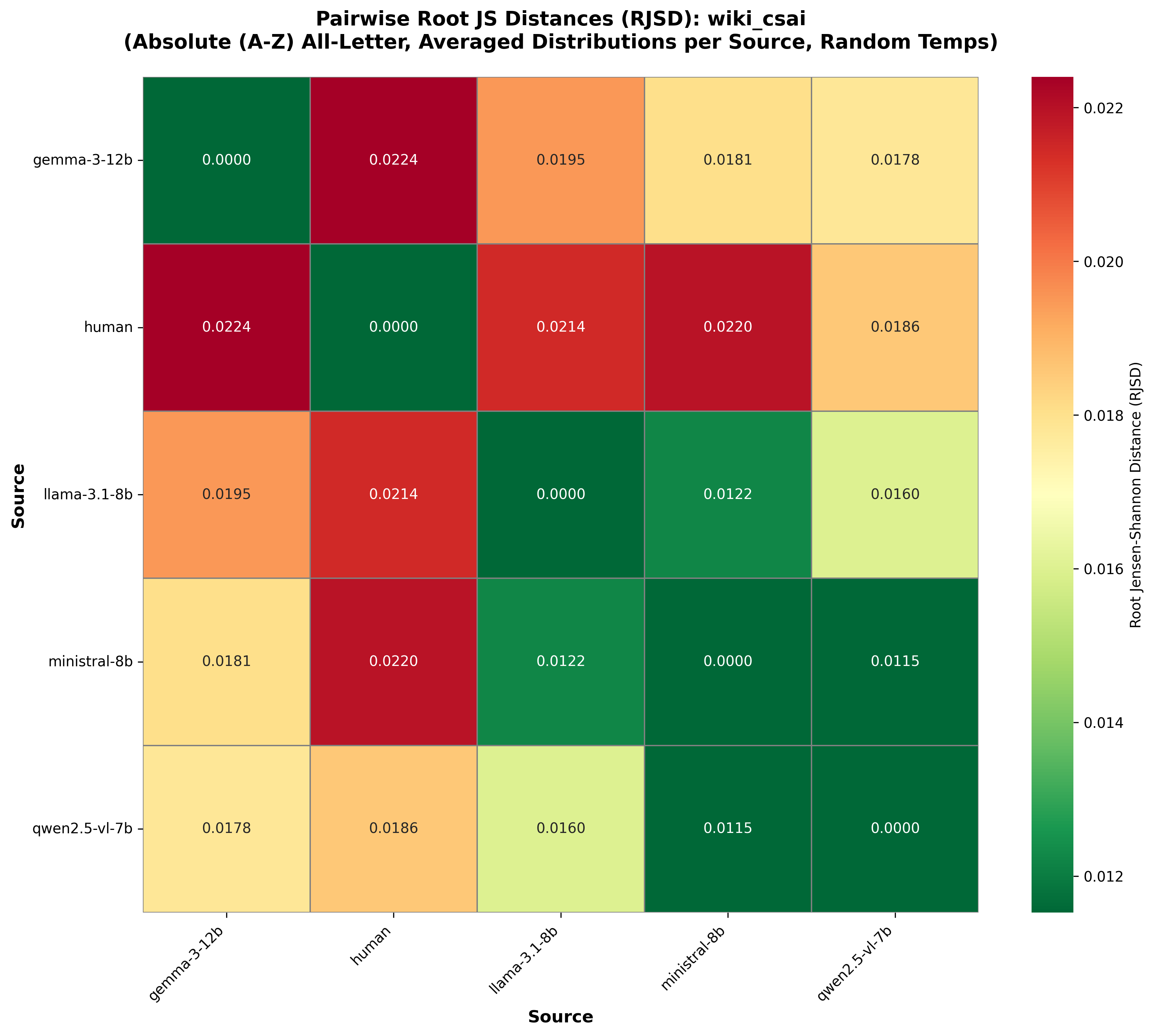}
        \caption{Wiki CSAI}
        \label{fig:mdt_wiki_pairwise}
    \end{subfigure}
    
    \vspace{0.5cm}
    
    \begin{subfigure}[b]{0.24\textwidth}
        \centering
        \includegraphics[width=\textwidth]{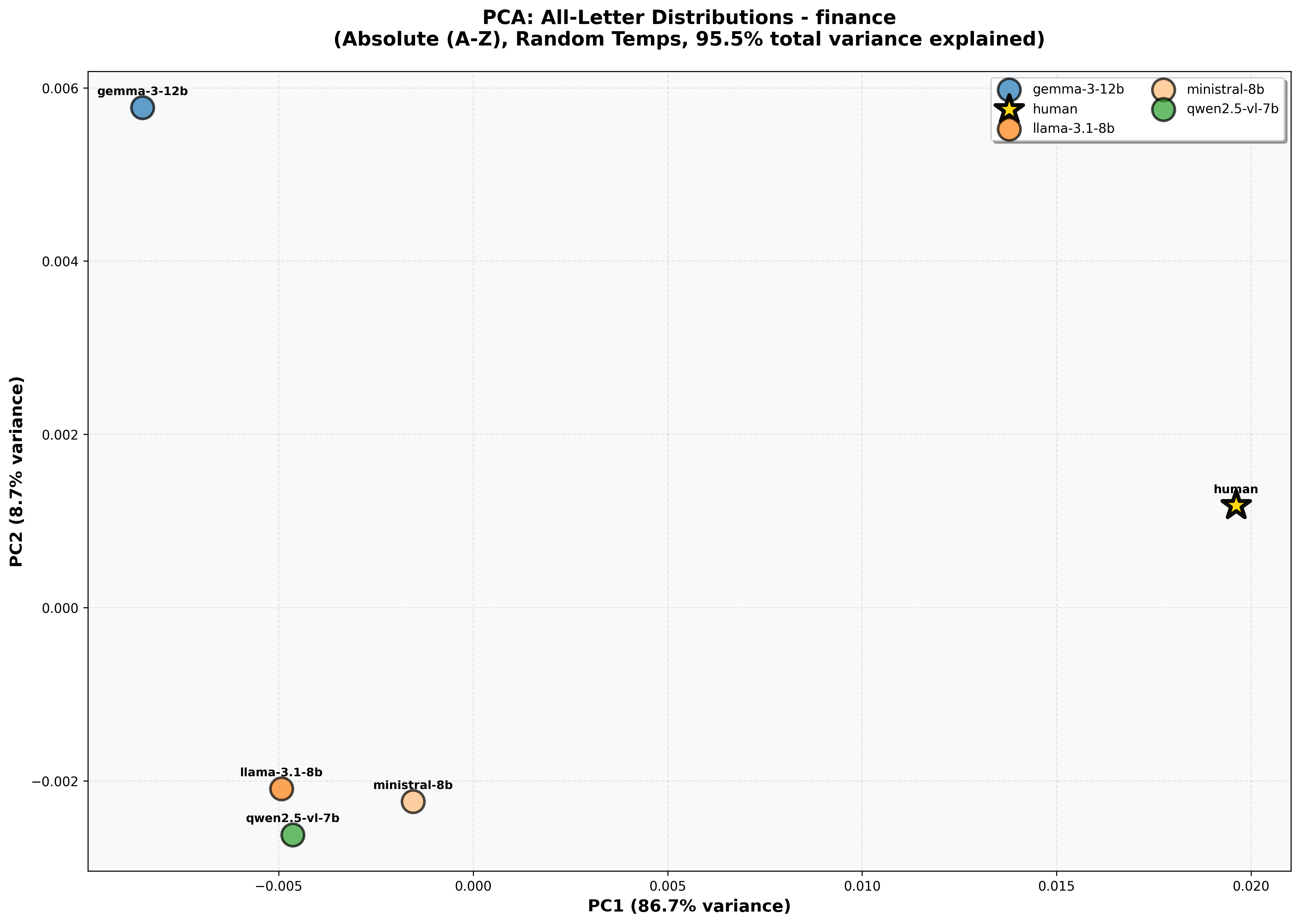}
        \caption{Finance}
        \label{fig:mdt_finance_pca}
    \end{subfigure}
    \hfill
    \begin{subfigure}[b]{0.24\textwidth}
        \centering
        \includegraphics[width=\textwidth]{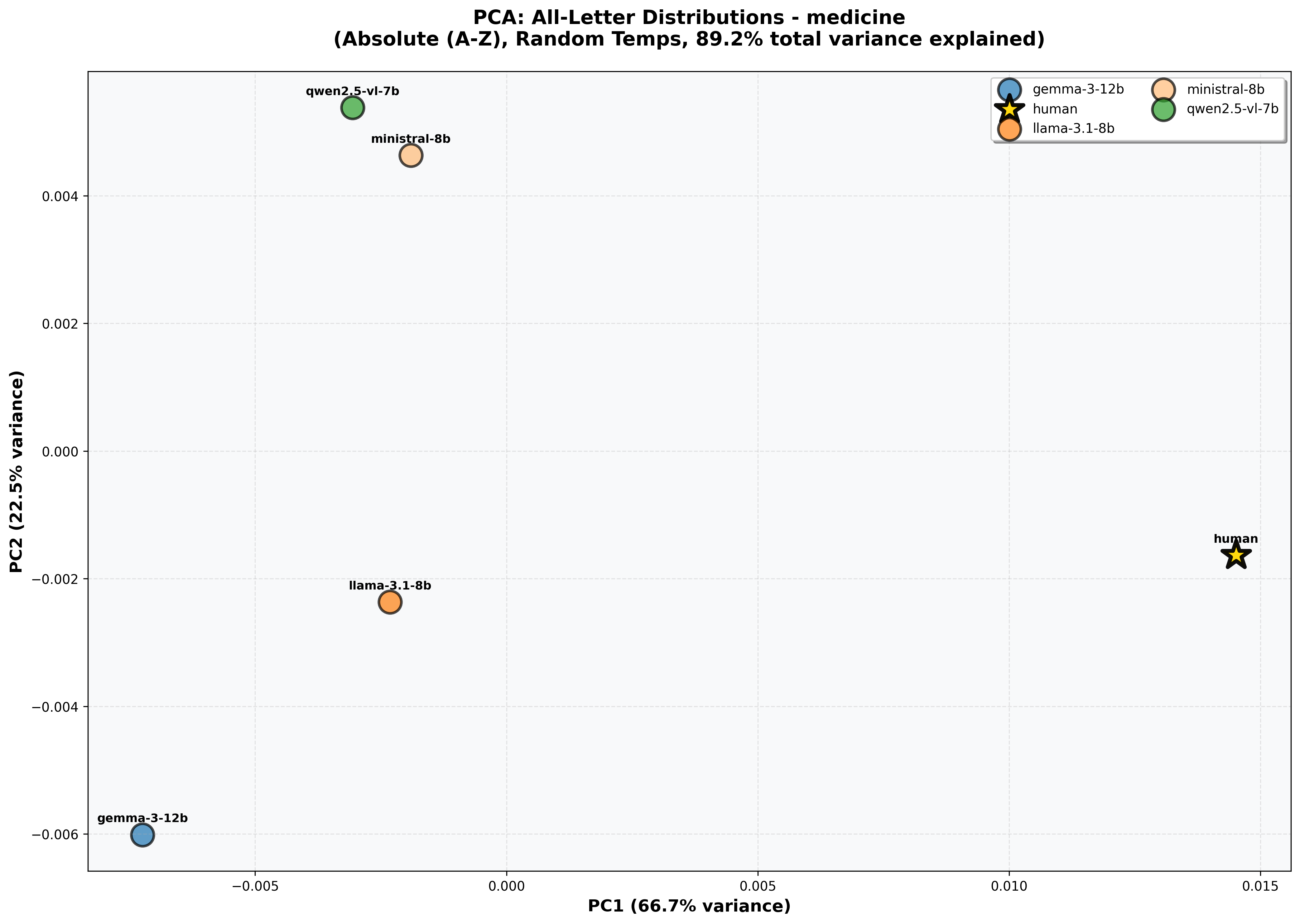}
        \caption{Medicine}
        \label{fig:mdt_medicine_pca}
    \end{subfigure}
    \hfill
    \begin{subfigure}[b]{0.24\textwidth}
        \centering
        \includegraphics[width=\textwidth]{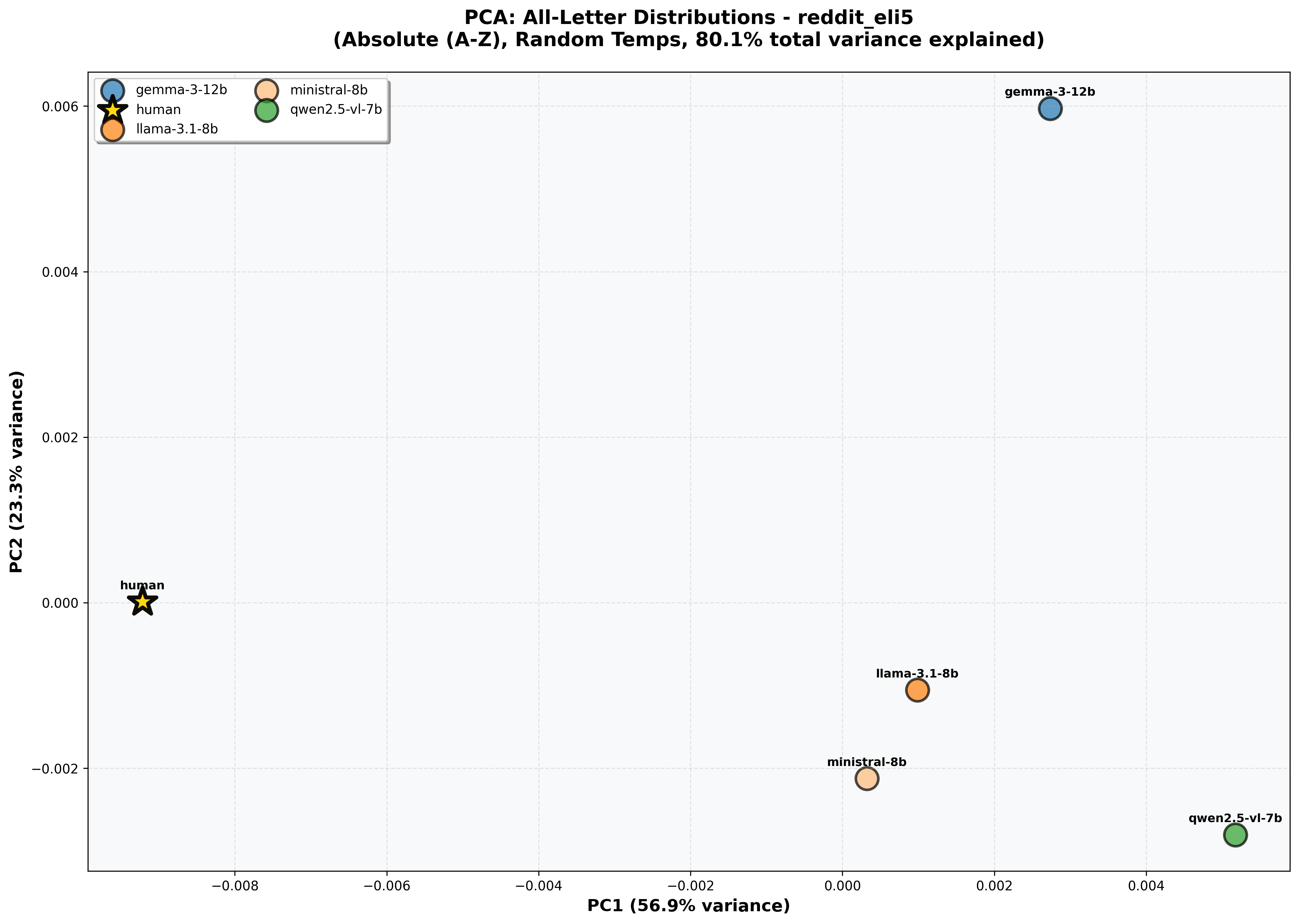}
        \caption{Reddit ELI5}
        \label{fig:mdt_reddit_pca}
    \end{subfigure}
    \hfill
    \begin{subfigure}[b]{0.24\textwidth}
        \centering
        \includegraphics[width=\textwidth]{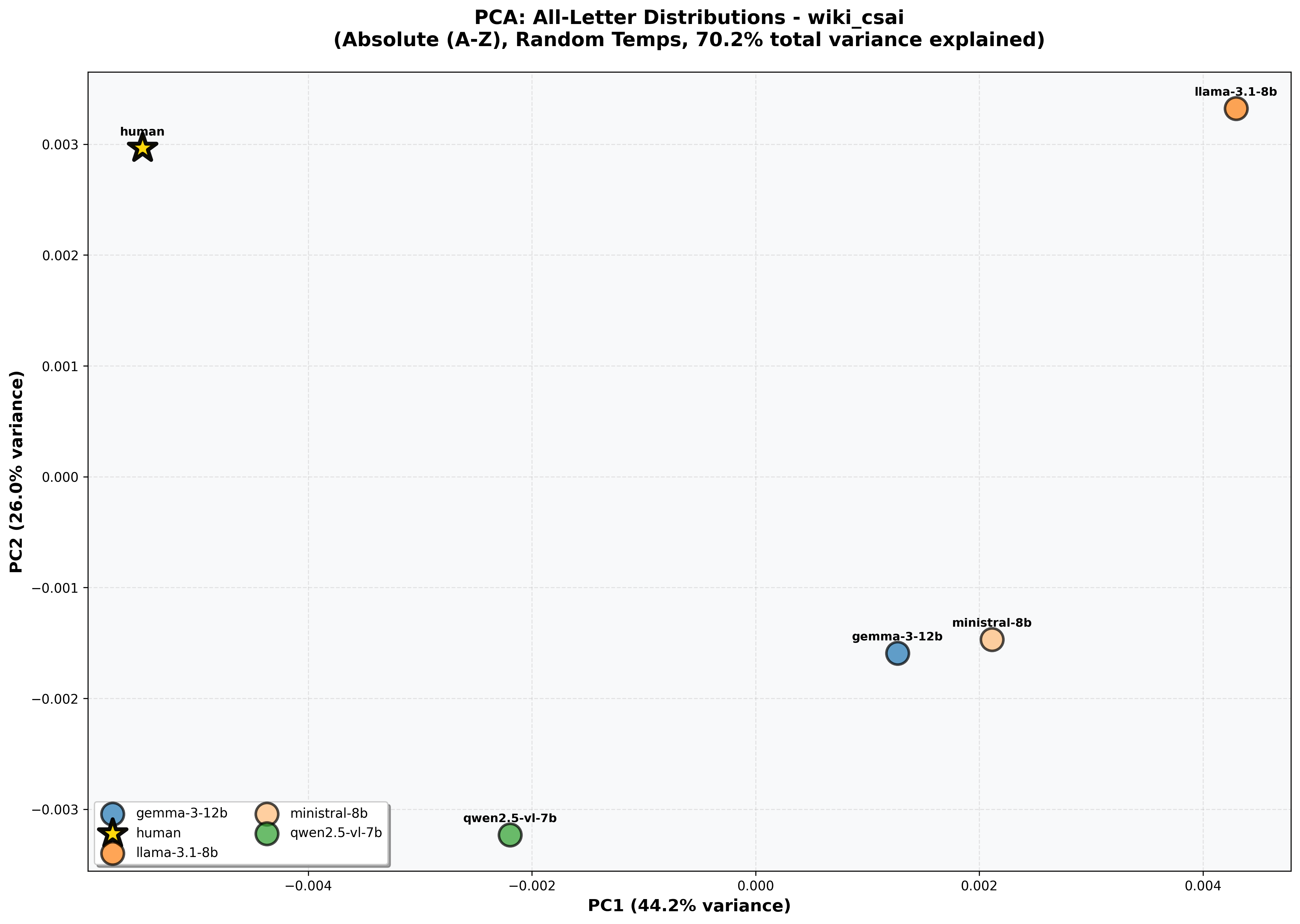}
        \caption{Wiki CSAI}
        \label{fig:mdt_wiki_pca}
    \end{subfigure}
    
    \vspace{0.5cm}
    
    \begin{subfigure}[b]{0.24\textwidth}
        \centering
        \includegraphics[width=\textwidth]{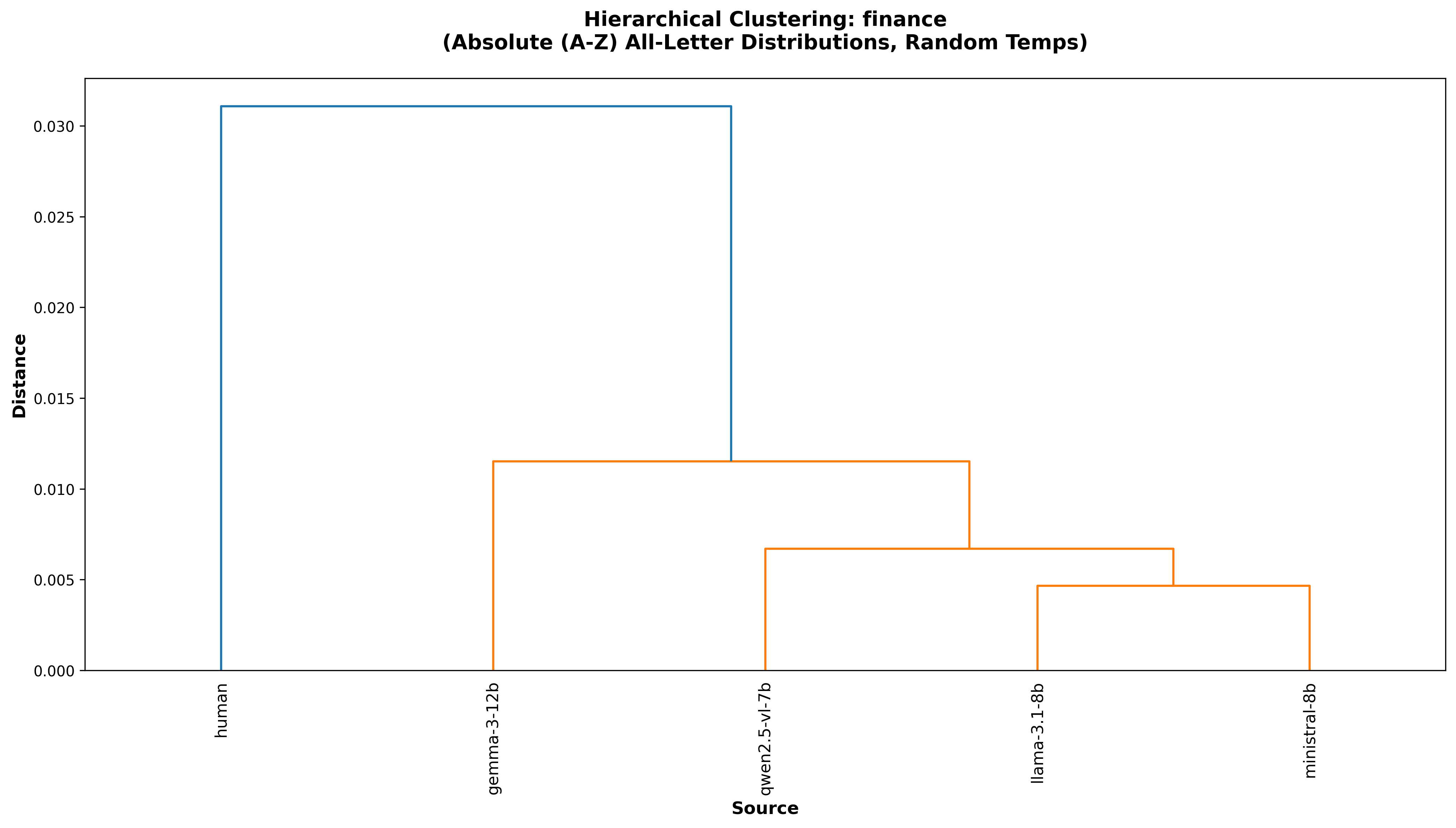}
        \caption{Finance}
        \label{fig:mdt_finance_dendrogram}
    \end{subfigure}
    \hfill
    \begin{subfigure}[b]{0.24\textwidth}
        \centering
        \includegraphics[width=\textwidth]{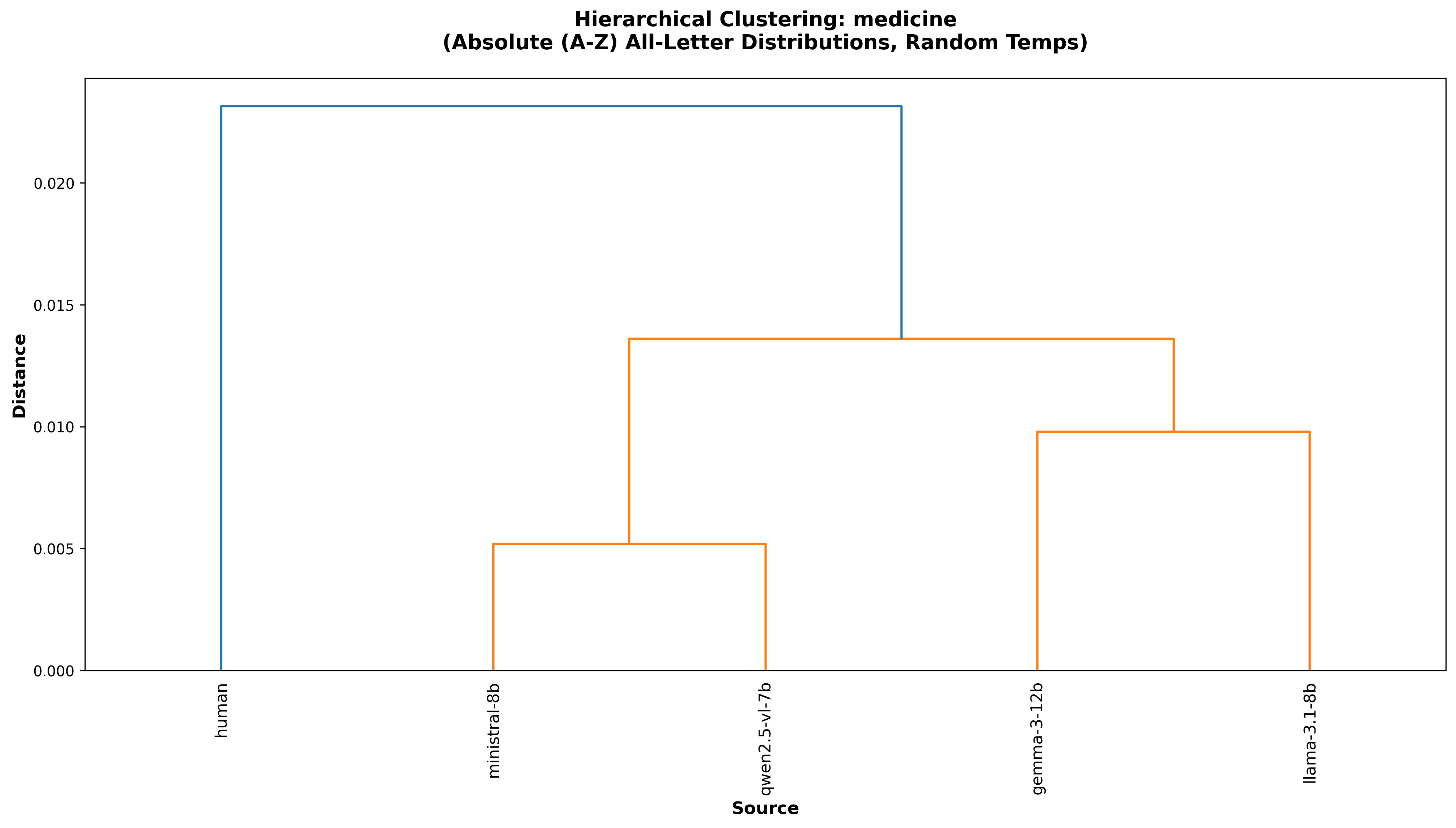}
        \caption{Medicine}
        \label{fig:mdt_medicine_dendrogram}
    \end{subfigure}
    \hfill
    \begin{subfigure}[b]{0.24\textwidth}
        \centering
        \includegraphics[width=\textwidth]{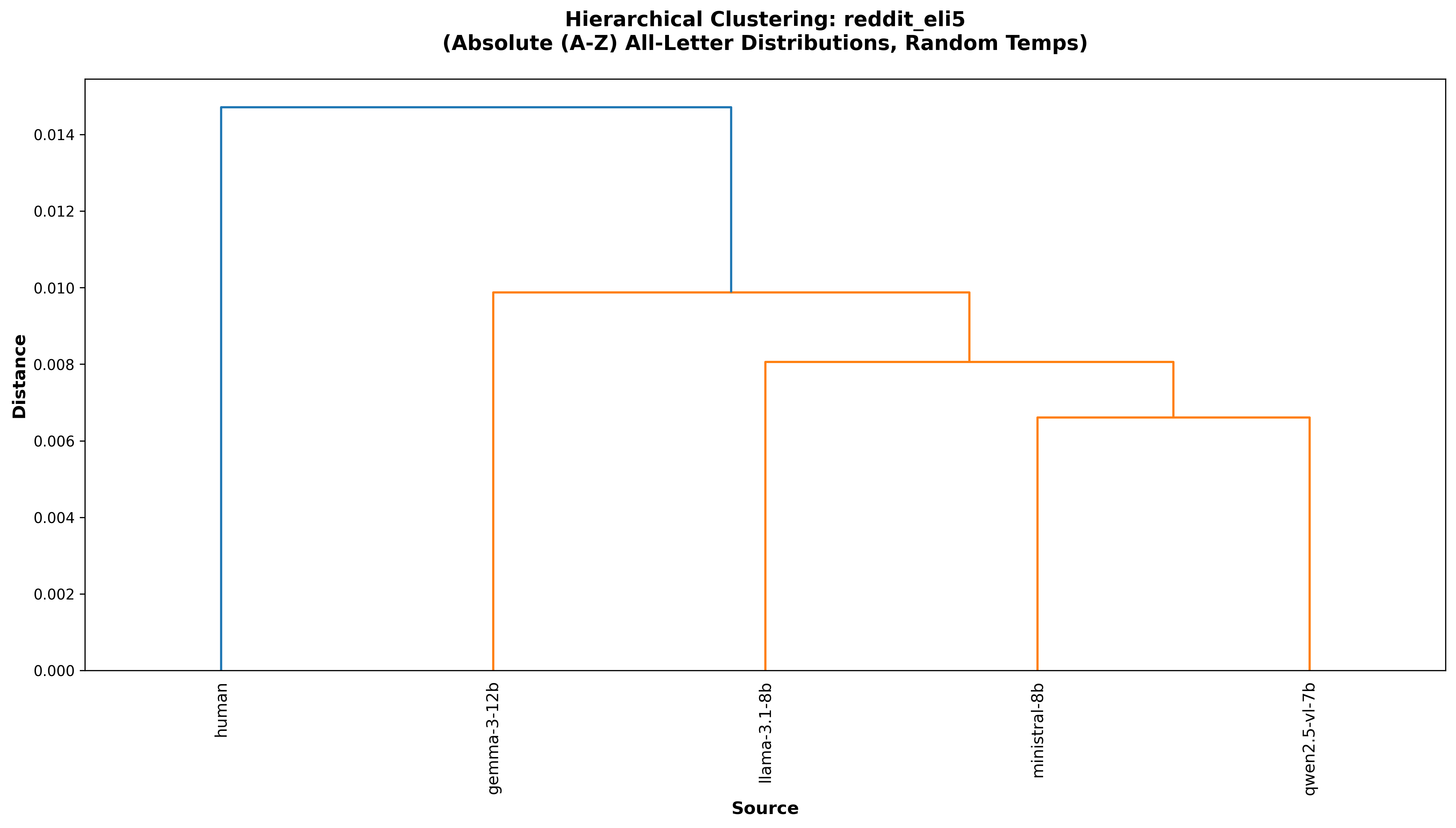}
        \caption{Reddit ELI5}
        \label{fig:mdt_reddit_dendrogram}
    \end{subfigure}
    \hfill
    \begin{subfigure}[b]{0.24\textwidth}
        \centering
        \includegraphics[width=\textwidth]{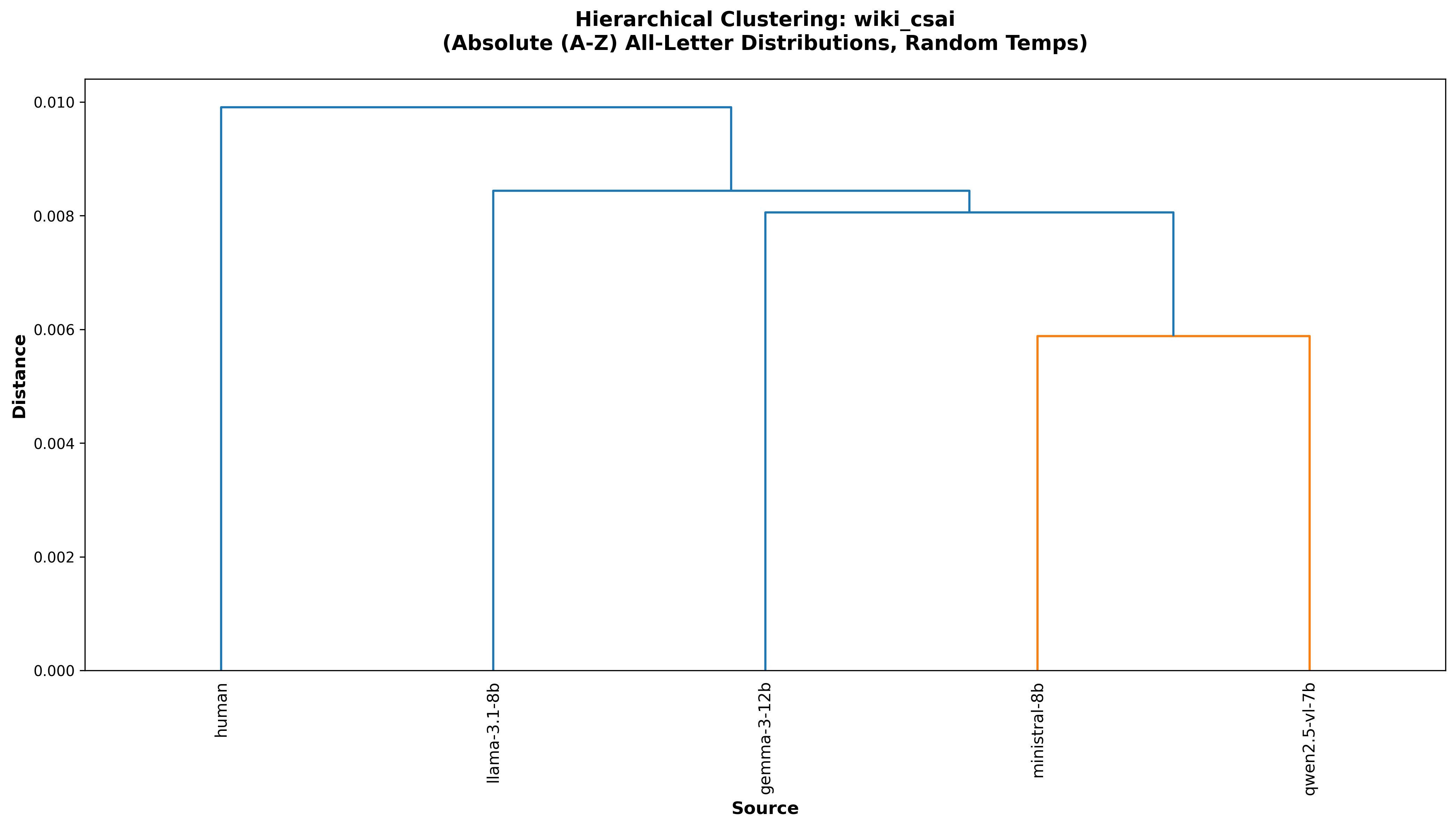}
        \caption{Wiki CSAI}
        \label{fig:mdt_wiki_dendrogram}
    \end{subfigure}
    
    \caption{Domain-specific LD-Score analysis on some domains of the MDTA dataset. Top row: Pairwise distance matrices. Middle row: PCA projections. Bottom row: Hierarchical clustering dendrograms. Finance and Medicine (specialized domains) exhibit stronger separation than Reddit ELI5 and Wiki CSAI (general knowledge), consistent with Eq. \ref{eq:total_divergence}'s domain bias prediction.}
    \label{fig:mdt_domain_analysis}
\end{figure}

Table~\ref{fig:log_probs} summarizes the composition of the entire MDTA dataset, with Table~\ref{tab:mdt_overview} detailing the average word and character counts per sample across domains.  Figure~\ref{fig:mdt_domain_analysis} 
presents domain-specific LD-Score analysis across these domains, extending our earlier 
Ghostbuster findings to a larger and more diverse benchmark with additional models and 
temperature variations.

\begin{table}[h]
\centering
\vspace{0.1cm}
\resizebox{\columnwidth}{!}{%
\begin{tabular}{lrrrrrrr}
\toprule
 & & \multicolumn{3}{c}{Standard} & \multicolumn{3}{c}{Adversarial} \\
\cmidrule(lr){3-5} \cmidrule(lr){6-8}
Domain & Human & $t{=}0.2$ & $t{=}0.5$ & $t{=}0.8$ & Para. & Avoid $\ell_{1}$ & Avoid $\ell_{1}$ \& $\ell_{2}$ \\
\midrule
Finance        & 3{,}933  & 15{,}732 & 15{,}732 & 15{,}732 & 15{,}732 & 15{,}732 & 15{,}732 \\
Medicine       & 1{,}248  & 4{,}992  & 4{,}992  & 4{,}992  & 4{,}992  & 4{,}992  & 4{,}992  \\
Open QA        & 1{,}187  & 4{,}748  & 4{,}748  & 4{,}748  & 4{,}748  & 4{,}748  & 4{,}748  \\
Reddit ELI5    & 51{,}336 & 68{,}448 & 68{,}448 & 68{,}448 & 68{,}448 & 68{,}448 & 68{,}448 \\
Wiki CSAI      & 842      & 3{,}368  & 3{,}368  & 3{,}368  & 3{,}368  & 3{,}368  & 3{,}368  \\
\midrule
\textbf{Total} & \textbf{58{,}546} & \textbf{97{,}288} & \textbf{97{,}288} & \textbf{97{,}288} & \textbf{97{,}288} & \textbf{97{,}288} & \textbf{97{,}288} \\
\bottomrule
\end{tabular}%
}
\caption{Dataset composition by domain in the MDTA benchmark. All AI models possess the same number of samples at each temperature. Adversarial variants (Paraphrase, Avoid $\ell_{1}$, Avoid $\ell_{1}$ \& $\ell_{2}$) are generated from the $t=0.5$ responses using the originating model. Reddit\_ELI5 has a particularly large number of human samples because the MDTA dataset contains 3 human responses per prompt.}
\label{tab:mdt-domain-composition}
\vspace{-0.2cm}
\end{table}

\begin{table}[h]
    \centering
    \caption{MDTA benchmark dataset statistics across domains.}
    \label{tab:mdt_overview}
    \begin{tabular}{lrrr}
        \toprule
        \textbf{Domain} & \textbf{Avg Words} & \textbf{Avg Chars} \\
        \midrule
        Finance      & 145 & 947   \\
        Medicine      & 191 & 1,120 \\
        Open QA       & 45  & 268   \\
        Reddit ELI5 & 160 & 891   \\
        Wiki CSAI     & 123 & 916   \\
        \bottomrule
    \end{tabular}
\end{table}

Figure~\ref{fig:mdt_domain_analysis} presents analogous analysis on the MDTA benchmark dataset, which includes additional models and temperature variations.

The MDTA results (Figure~\ref{fig:mdt_domain_analysis}) corroborate findings from the Ghostbuster dataset. Hierarchical clustering consistently shows human text splitting at the top level across all domains. Similar models cluster together within the AI group, with the dendrogram structure reflecting training data overlap. Notably, specialized domains (Finance, Medicine, Reddit ELI5) display stronger separation and clearer hierarchical structure compared to general domains (Wiki CSAI), validating that domain specialization amplifies the detection signal as predicted by our theoretical framework.

\subsection{Adversarial Dataset Analysis (MDTA)}

\begin{figure}[t]
    \centering
    \includegraphics[width=1\linewidth]{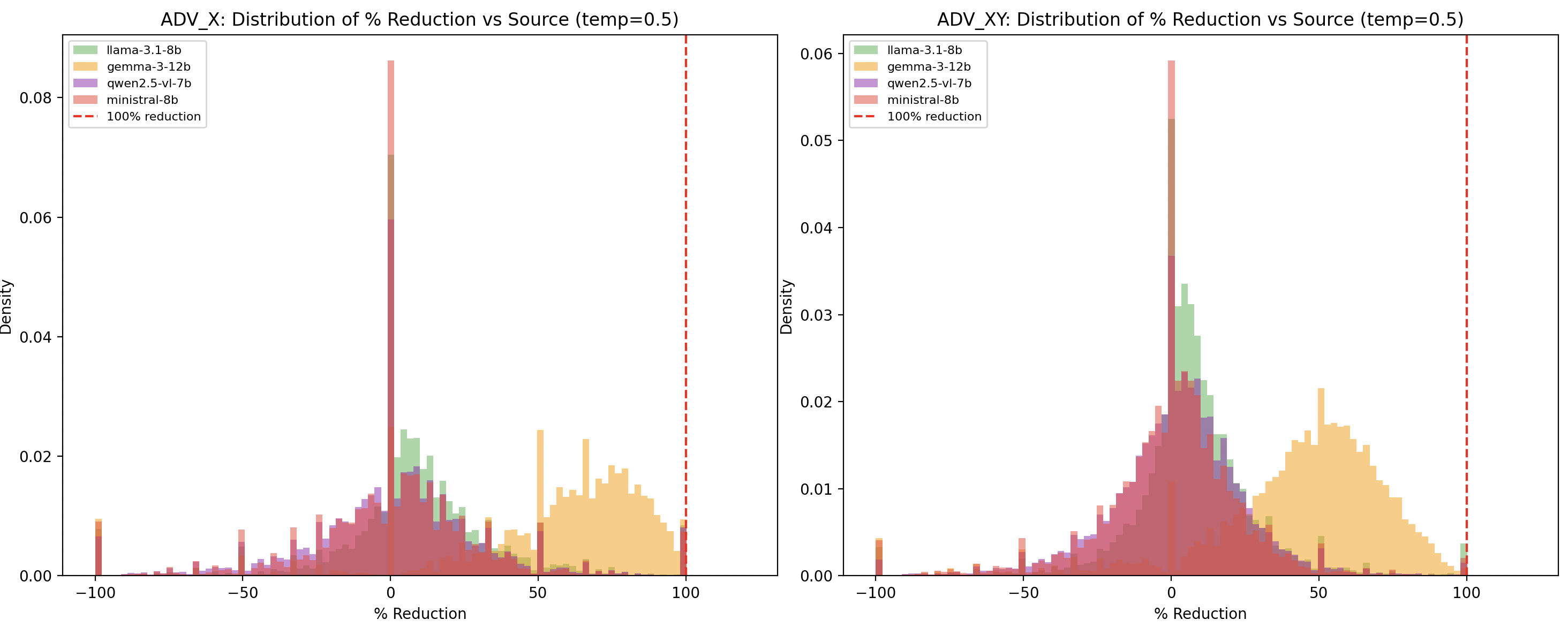}
    \caption{Distribution of percentage reduction in target letter frequency for the \texttt{ADV\_X} (single-letter removal) and \texttt{ADV\_XY} (two-letter removal) adversarial attacks,      
  aggregated across all samples and domains in the MDTA dataset. The $x$-axis shows the percentage reduction in occurrences of the target letter(s) in the adversarial output relative to the
  original model response ($\text{temp} = 0.5$). A value of $100\%$ (red dashed line) indicates complete removal of the target letter(s). Negative values indicate that the attack             
  inadvertently \textit{increased} the frequency of the target letter(s) in the output. \texttt{gemma-3-12b} achieves the highest reduction rates, while \texttt{ministral-8b} and             
  \texttt{qwen2.5-vl-7b} frequently fail to reduce---or even add---target letters.}                                                                                  
    \label{fig:percent_reduction}
\end{figure}

\begin{figure}[t]
    \centering
    \includegraphics[width=\linewidth]{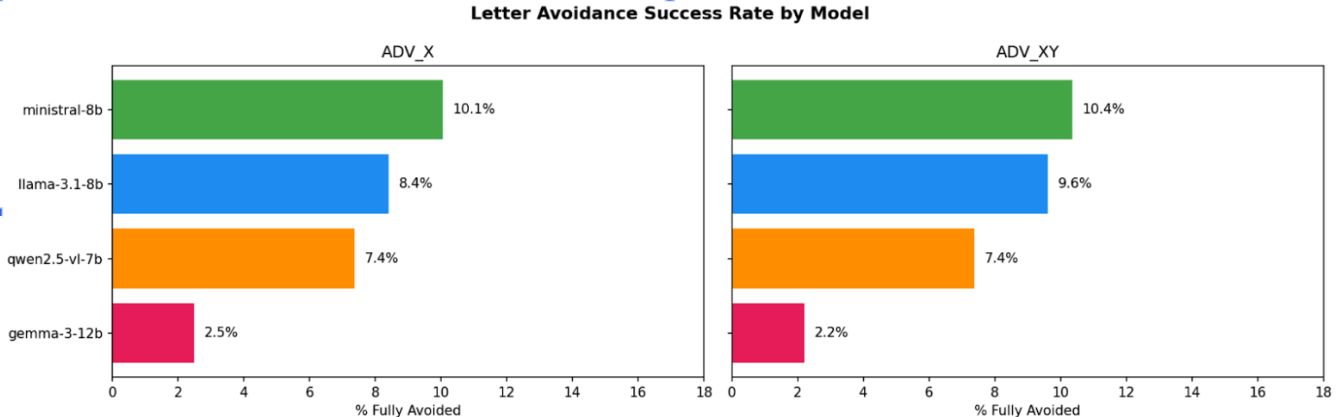}
    \caption{Percentage of samples where target letter(s) were fully absent from model outputs,
    averaged across all target letters, under single-letter (\textsc{ADV\_X}) and two-letter
    (\textsc{ADV\_XY}) constraints.}
    \label{fig:avoidance_model}
\end{figure}

\begin{figure}[t]
    \centering
    \includegraphics[width=\linewidth]{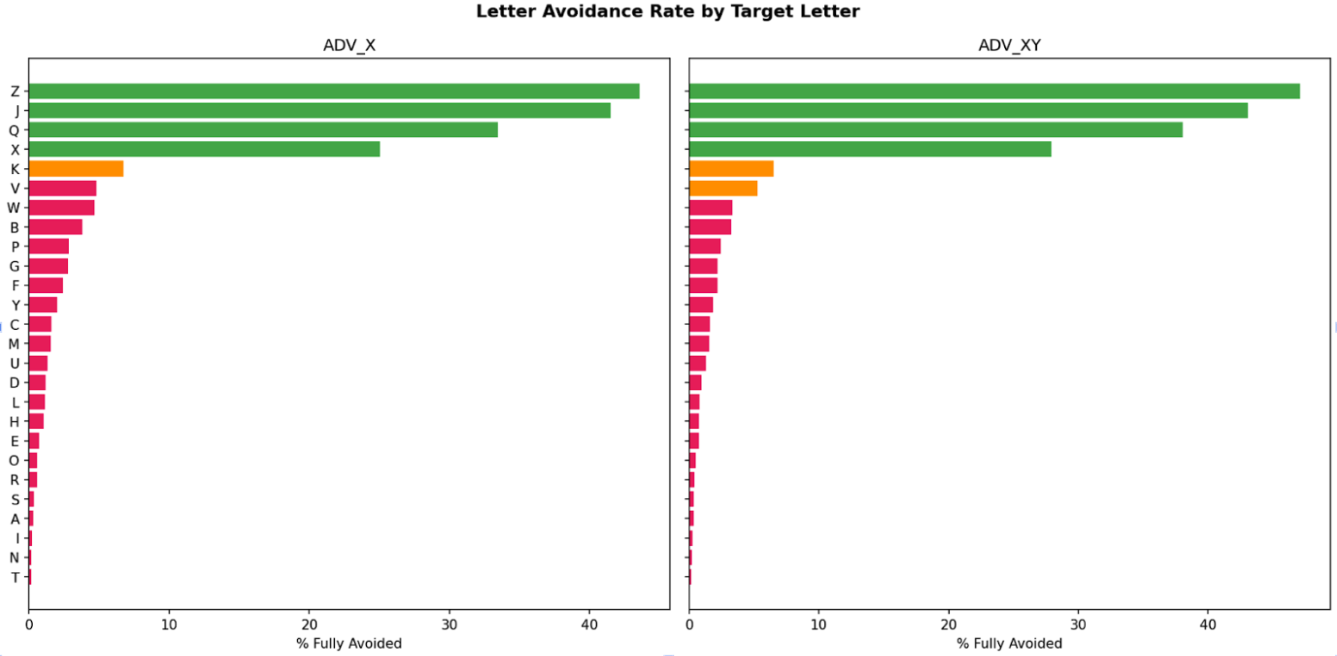}
    \caption{Per-letter avoidance success rate (100\% removal) under \textsc{ADV\_X} and \textsc{ADV\_XY},
    sorted in descending order.}
    \label{fig:avoidance_letter}
\end{figure}

As mentioned earlier, for each AI-generated response at temperature 0.5, we used the originating model itself to produce three adversarial rewrites: (A) a standard paraphrase, (B) a paraphrase avoiding a randomly
selected letter $\ell_1$, and (C) a paraphrase simultaneously avoiding two letters $\ell_1 \neq \ell_2$.
We perform analysis. In Figures~\ref{fig:percent_reduction}~\ref{fig:avoidance_model}~\ref{fig:avoidance_letter}, we analyze the effectiveness of letter-removal attacks (B) and (C) in altering the responses of the models.
 
Overall, 100\% avoidance rates are low across all models (Figure~\ref{fig:avoidance_model}), with the
best-performing model (\texttt{ministral-8b}) achieving full avoidance in only $\sim$10\% of samples.
As shown in Figure~\ref{fig:avoidance_letter}, success is heavily skewed toward rare letters---\texttt{Z},
\texttt{J}, \texttt{Q}, and \texttt{X} are avoided in 30--45\% of samples, while high-frequency
letters like \texttt{T}, \texttt{S}, and \texttt{N} are nearly impossible to avoid. This means
adversarial character-distribution manipulation succeeds precisely where it matters least for
detection: rare letters contribute little to the LD-Score's discriminative signal, leaving the
detector's core features largely intact.
\subsection{Linguistic Complexity Analysis}
We assess text complexity through readability, lexical diversity, and $n$-gram analysis.

\textbf{$n$-gram Analysis.} We further perform $n$-gram analysis by examining frequency distributions of contiguous word sequences, where an $n$-gram consists (uni and bi) of $n$ consecutive words. According to Table~\ref{tab:ngram-aggregated}, it is clearly evident that the cumulative human unigram-vocabulary is higher than any of the AI models. This is in accordance to Subsection ~\ref{sec:theory-foundation} (Convergence through Exposure) since the human responses are sourced from various different speakers.

\begin{figure}[t]
    \centering
    \includegraphics[width=1\linewidth]{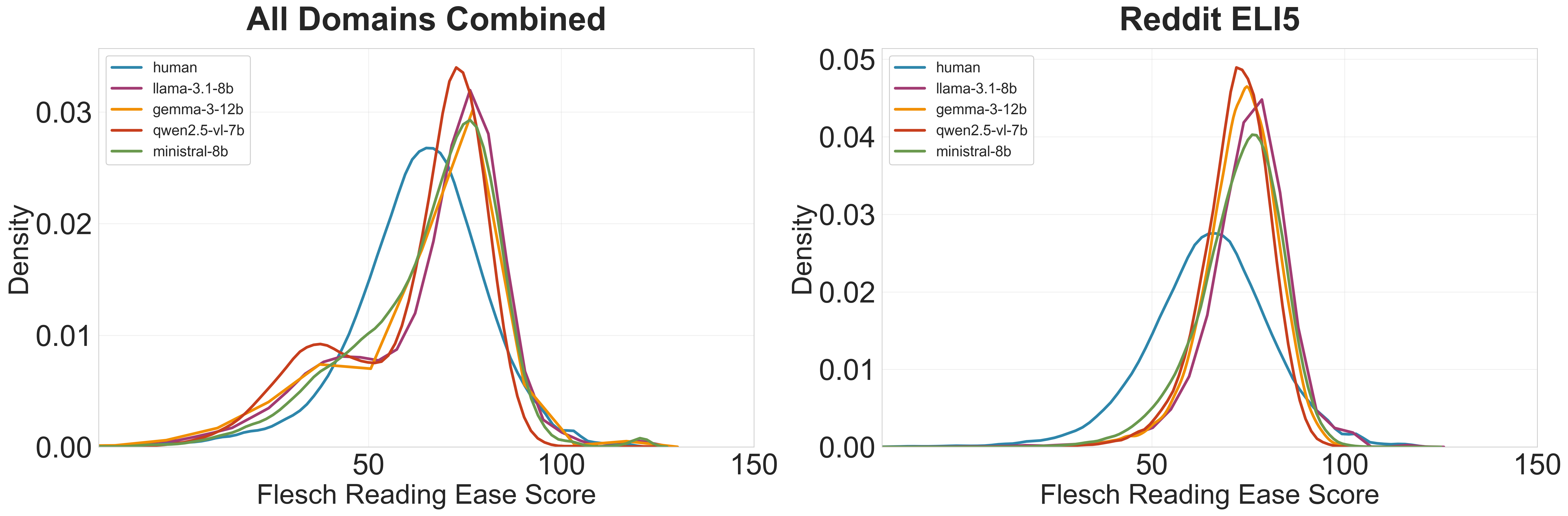}
    \caption{Instance-level readability analysis  (FKGL Kernel Density Estimation (KDE)) across domains shows that human-written text is consistently simpler than model-generated text, with the gap most pronounced in the Reddit\_ELI5 domain.}
    \label{fig:readability}
\end{figure}

\textbf{Readability.} Readability is quantified using the Flesch--Kincaid Grade Level (FKGL) \cite{flesch1948new}, defined as
\begin{equation}
\mathrm{FKGL}
= 0.39 \cdot \frac{W}{S}
+ 11.8 \cdot \frac{Sy}{W}
- 15.59,
\end{equation}
where $W$ denotes the number of words, $S$ the number of sentences, and $Sy$ the number of syllables. FKGL estimates the U.S. grade level required to comprehend the text. The results, as depicted in Fig.~\ref{fig:readability}, are expected. A specific human tends to explain simply, using smaller and relatively simpler words than language models. This is amplified if we only visualize the Reddit\_ELI5 domain. While $n$-gram analysis is performed considering the entire human corpus within the dataset, readability analysis is drawn per sample.

\begin{figure}[t]
    \centering
    \includegraphics[width=1\linewidth]{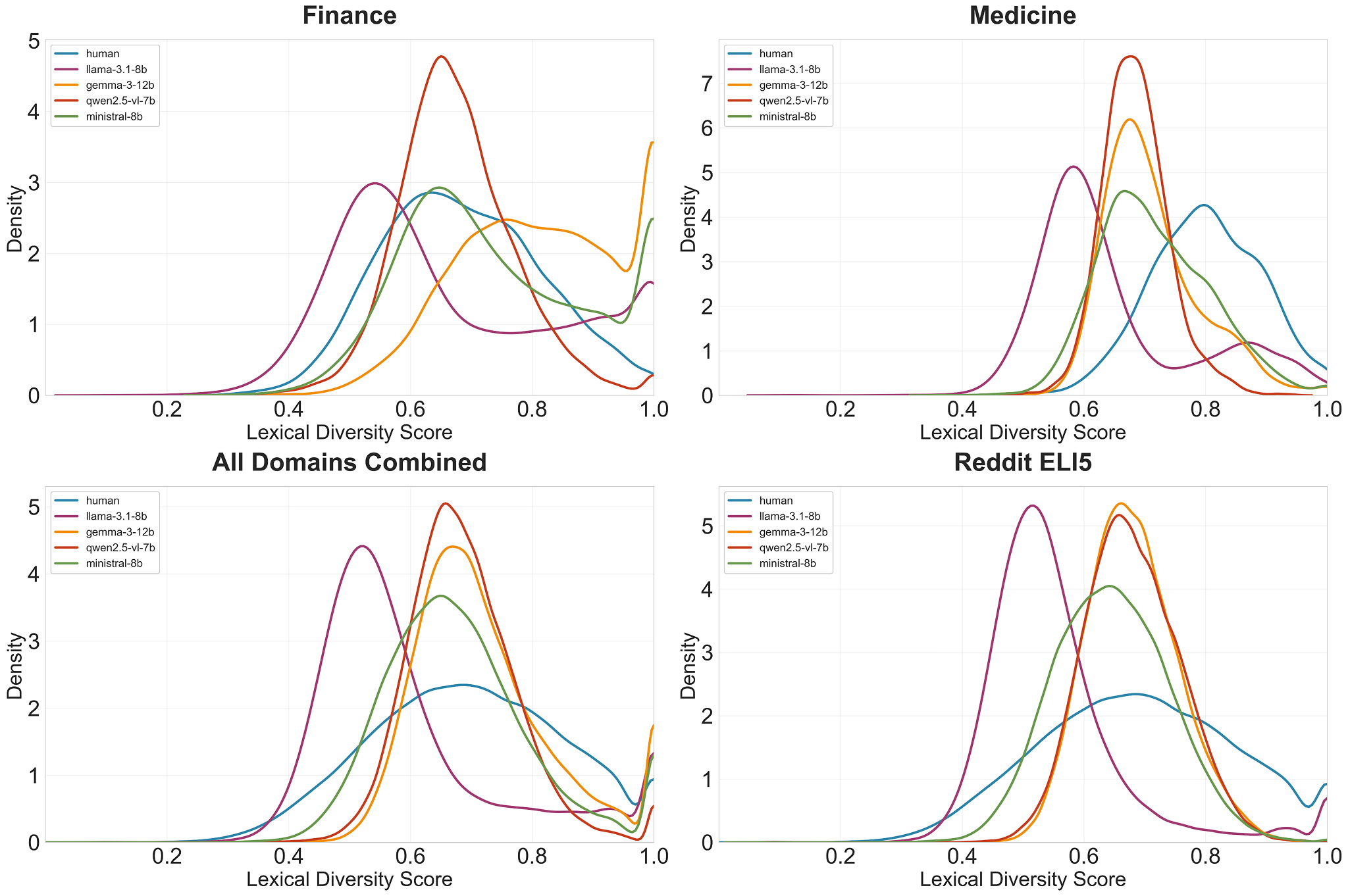}
    \caption{Instance-level lexical diversity (LDS) distributions across domains (Finance, Medicine, Reddit ELI5, and aggregated). Within specialized domains, human-written text exhibits LDS closer to those of AI-generated text, particularly in Finance and Medicine, while Reddit\_ELI5 shows substantially higher LDS-variability between humans and AI.}
    \label{fig:lexical_diversity}
\end{figure}

\textbf{Lexical Diversity.} We calculate the lexical diversity score (LDS) as defined in \cite{wu2025detectrlbenchmarkingllmgeneratedtext}.
\begin{equation}
\mathrm{LDS} = \frac{|V|}{N},
\end{equation}
where $|V|$ is the number of unique word types and $N$ is the total number of word in the input text. Higher LDS values indicate richer vocabulary usage, while lower values suggest more repetitive language. We observe similar results: a particular human uses simpler words, while language models use complicated words with much less repetition. Fig.~\ref{fig:lexical_diversity} shows the analysis results calculated per sample. 

To facilitate reproducibility and further research in model-agnostic AI text detection, the dataset will be publicly released upon acceptance.

\begin{table}[t]

\vspace{0.1cm}
\centering
\begin{tabular}{lrr}
\toprule
Source & Unigrams & Bigrams \\
\midrule
Human         & 72{,}468 & 1{,}476{,}050 \\
LLaMA-3.1-8B  & 50{,}484 & 1{,}377{,}721 \\
Gemma-3-12B   & 43{,}059 & 1{,}299{,}946 \\
Qwen2.5-VL-7B & 45{,}167 & 1{,}369{,}694 \\
Ministral-8B  & 42{,}824 & 1{,}078{,}912 \\
\bottomrule
\end{tabular}
\caption{$N$-gram counts across all domains in the MDTA dataset separated by source, calculated by concatenating all samples.}
\label{tab:ngram-aggregated}
\vspace{-0.2cm}
\end{table}

\subsection{Domain Specialization Effects Summary}

The domain-specific results validate our theoretical prediction (Eq. \ref{eq:total_divergence}) that the magnitude of human-AI separation varies with domain specialization level. Specialized domains (Reuters, Finance, Medicine, Reddit\_ELI5) with structured, technical vocabulary/way of writing exhibit the strongest separation. General-purpose domains (Creative Writing, Wiki CSAI) with diverse and unstructured vocabulary closer to $P_{\text{global}}(w)$ show diminished separation. When pooled across all domains, the clustering inequality tightens, but the Wall of Separation persists, confirming the robustness of letter distribution signatures for AI text detection.

\clearpage
\section{More Experiment Results}
\label{sec:more-exp}

\subsection{Ghostbuster Experiment Result}

\begin{table}[h]
\centering
\resizebox{0.8\textwidth}{!}{%
\begin{tabular}{l | ccc | c}
\toprule
\textbf{Method} & Essay & Reuters & WP & Avg \\
\midrule
\multicolumn{5}{l}{\textit{AUROC}} \\
\midrule
DNA & \textbf{1.000}\tiny{$\pm$0.000} & 0.999{\tiny{$\pm$0.003}} & 0.984{\tiny{$\pm$0.009}} & 0.994{\tiny{$\pm$0.005}} \\
Bino & \textbf{1.000}\tiny{$\pm$0.001} & 0.995{\tiny{$\pm$0.005}} & 0.961{\tiny{$\pm$0.012}} & 0.985{\tiny{$\pm$0.008}} \\
FDGPT & 0.977{\tiny{$\pm$0.013}} & 0.962{\tiny{$\pm$0.016}} & 0.865{\tiny{$\pm$0.026}} & 0.935{\tiny{$\pm$0.019}} \\
\midrule
LD-DNA & \textbf{1.000}\tiny{$\pm$0.000} & \textbf{1.000}\tiny{$\pm$0.001} & 0.985{\tiny{$\pm$0.008}} & \textbf{0.995}\tiny{$\pm$0.005} \\
LD-Bino & \textbf{1.000}\tiny{$\pm$0.000} & 0.999{\tiny{$\pm$0.002}} & 0.967{\tiny{$\pm$0.008}} & 0.989{\tiny{$\pm$0.005}} \\
LD-FDGPT & 0.992{\tiny{$\pm$0.005}} & 0.994{\tiny{$\pm$0.005}} & 0.893{\tiny{$\pm$0.027}} & 0.960{\tiny{$\pm$0.016}} \\
\midrule
DNA+Bino & \textbf{1.000}\tiny{$\pm$0.000} & 0.999{\tiny{$\pm$0.001}} & 0.981{\tiny{$\pm$0.008}} & 0.993{\tiny{$\pm$0.005}} \\
DNA+FDGPT & \textbf{1.000}\tiny{$\pm$0.001} & 0.999{\tiny{$\pm$0.001}} & \textbf{0.986}\tiny{$\pm$0.006} & \textbf{0.995}\tiny{$\pm$0.004} \\
\midrule
\multicolumn{5}{l}{\textit{F1 Score}} \\
\midrule
DNA & 0.991{\tiny{$\pm$0.006}} & 0.978{\tiny{$\pm$0.012}} & 0.945{\tiny{$\pm$0.021}} & 0.971{\tiny{$\pm$0.014}} \\
Bino & 0.981{\tiny{$\pm$0.018}} & 0.952{\tiny{$\pm$0.030}} & 0.895{\tiny{$\pm$0.032}} & 0.943{\tiny{$\pm$0.027}} \\
FDGPT & 0.918{\tiny{$\pm$0.024}} & 0.893{\tiny{$\pm$0.025}} & 0.780{\tiny{$\pm$0.044}} & 0.864{\tiny{$\pm$0.032}} \\
\midrule
LD-DNA & \textbf{0.996}\tiny{$\pm$0.004} & \textbf{0.996}\tiny{$\pm$0.004} & \textbf{0.949}\tiny{$\pm$0.011} & \textbf{0.980}\tiny{$\pm$0.007} \\
LD-Bino & 0.990{\tiny{$\pm$0.005}} & 0.988{\tiny{$\pm$0.006}} & 0.918{\tiny{$\pm$0.021}} & 0.965{\tiny{$\pm$0.013}} \\
LD-FDGPT & 0.965{\tiny{$\pm$0.009}} & 0.974{\tiny{$\pm$0.009}} & 0.810{\tiny{$\pm$0.036}} & 0.916{\tiny{$\pm$0.022}} \\
\midrule
DNA+Bino & 0.992{\tiny{$\pm$0.005}} & 0.986{\tiny{$\pm$0.008}} & 0.935{\tiny{$\pm$0.011}} & 0.971{\tiny{$\pm$0.008}} \\
DNA+FDGPT & 0.994{\tiny{$\pm$0.006}} & 0.990{\tiny{$\pm$0.007}} & 0.943{\tiny{$\pm$0.014}} & 0.976{\tiny{$\pm$0.010}} \\
\bottomrule
\end{tabular}%
}
\caption{Detection performance on the Ghostbuster benchmark (Essay, Reuters, WP domains) with 100 balanced training samples. LD-X denotes augmentation of method X with the LD-Score via RBF-SVM fusion. Results are mean $\pm$ std over 5 runs (different seeds). \textbf{Bold} indicates best per column.}
\label{tab:ghostbuster_results}
\end{table}

\subsection{Results Based on Temperature.}
\label{sec:temp-results}
\begin{figure*}[h]
    \centering
    \includegraphics[width=1\linewidth]{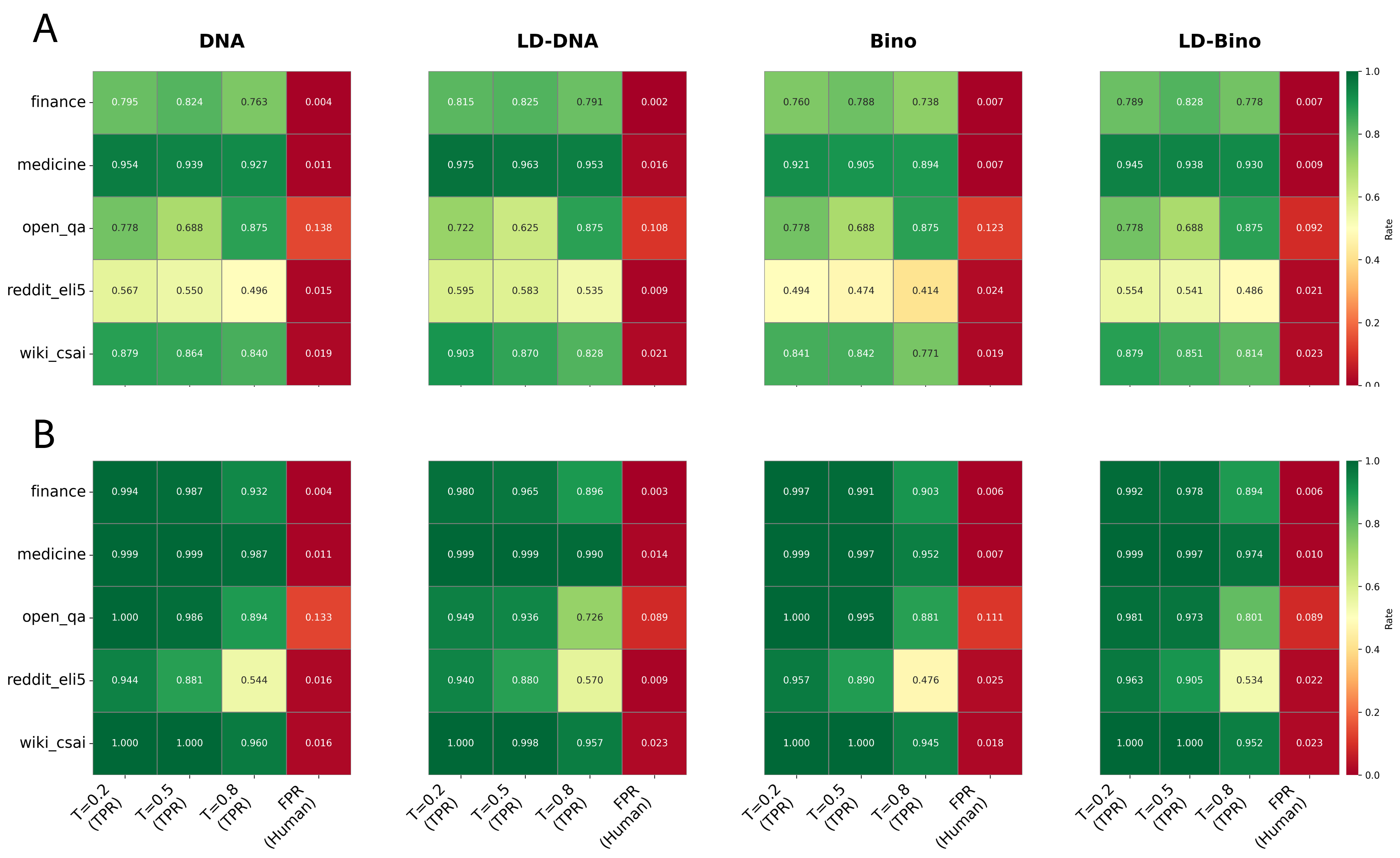}
    \caption{Temperature-dependent detection performance across domains for Gemma-3-12B \textbf{(A)} and Qwen2.5-VL-7B \textbf{(B)} models. Heatmaps show True Positive Rate (TPR) at different generation temperatures (T=0.2, 0.5, 0.8) and False Positive Rate (FPR) on human text across five domains and four detection methods. All methods use globally optimized thresholds. Increasing the temperature typically worsens the TPR. Augmenting with our approach mostly increase the TPR and reduces FPR, although there are exceptions.}
    \label{fig:temp-heatmap}
\end{figure*}

The MDTA dataset also provides model responses generated at different temperature settings. We compare baseline and augmented methods across temperatures by running the analysis on the same human samples while replacing only the AI-generated inputs with responses from different temperatures. As a result, the False Positive Rate (FPR) remains constant. The results are summarized in Fig.~\ref{fig:temp-heatmap}. Our augmented approaches generally yield higher TPR with lower FPR. As expected, TPR typically decreases as the generation temperature increases.

As temperature increases, AI text exhibits greater word-level diversity, which degrades raw perplexity signals. However, this same variability moves AI text further from human letter distributions on average. Thus, while perplexity alone worsens, perplexity + lexical/LD signals benefit from temperature-induced diversity, increasing TPR and often lowering FPR by amplifying the global-vs-domain distribution gap.

\subsection{Comparison with Stylometric Features}
\label{sec:stylo}

Recent work has explored rich stylometric feature sets for AI text detection. \citet{Przystalski_2026} construct a suite of 196 linguistically-motivated features spanning lexical diversity, syntactic complexity, and punctuation patterns, identifying the most discriminative subset via SHAP analysis on a multiclass attribution task. We compare LD-score against their top-10 SHAP-ranked features, evaluated jointly across the MDTA and Ghostbuster datasets (125 human / 125 AI per domain).

\paragraph{Results.} Table~\ref{tab:stylometry_snippet} reports F1 scores across all domains. LD-Score achieves the highest average F1 
(0.76), outperforming all individual stylometric features. Gains are most pronounced in 
structured domains such as Reuters (0.93 vs.\ 0.78), Medicine (0.87 vs.\ 0.79), and Essay 
(0.84 vs.\ 0.71). This is consistent with our theoretical prediction that domain-specialized 
vocabularies amplify the letter-distribution signal.

\paragraph{Fragility of stylometric features.} Despite the richness of the 196-feature suite, the highest-ranked features by SHAP importance are overwhelmingly simple surface-level statistics: punctuation frequency, comma rate, period count, words-per-sentence, and numeral density. These are precisely the features most vulnerable to adversarial manipulation-a model can trivially shift punctuation density or sentence length without altering semantic content. Shifting letter-level distributions, by contrast, requires coordinated vocabulary-level changes across the entire text. The dominance of surface features in the top-10 thus reflects a broader limitation of stylometry: in-distribution discriminability does not imply robustness under adaptive attack.

\begin{table}[h]
\centering

\resizebox{\textwidth}{!}{%
\begin{tabular}{lcccccccccc}
\toprule
\textbf{Feature} & \textbf{Essay} & \textbf{Finance} & \textbf{Medicine} & \textbf{Open QA} & \textbf{Reddit ELI5} & \textbf{Reuters} & \textbf{Wiki CSAI} & \textbf{WP} & \textbf{Avg} \\
\midrule
Pooled RJSD  & \textbf{0.84} & 0.66 & \textbf{0.87} & 0.66 & \textbf{0.78} & \textbf{0.93} & 0.66 & 0.71 & \textbf{0.76} \\
TTR Lemmas   & 0.71 & 0.67 & 0.76 & 0.67 & 0.66 & 0.74 & 0.67 & 0.77 & 0.71 \\
Commas       & 0.67 & 0.67 & 0.74 & 0.66 & 0.71 & 0.70 & 0.67 & \textbf{0.78} & 0.70 \\
Punctuation  & 0.66 & \textbf{0.77} & 0.79 & 0.67 & 0.69 & 0.67 & 0.66 & 0.67 & 0.70 \\
Noun Phrase  & 0.66 & 0.72 & 0.76 & 0.66 & 0.72 & 0.67 & 0.66 & 0.72 & 0.70 \\
Numerals     & 0.78 & 0.66 & 0.67 & 0.66 & 0.67 & 0.79 & \textbf{0.68} & 0.69 & 0.70 \\
Words/Sent   & 0.67 & 0.66 & 0.67 & \textbf{0.70} & 0.70 & 0.78 & 0.66 & 0.72 & 0.70 \\
Sent Diff    & 0.67 & 0.67 & 0.67 & 0.65 & 0.76 & 0.77 & 0.66 & 0.71 & 0.69 \\
Dots         & 0.66 & 0.65 & 0.71 & 0.69 & 0.67 & 0.68 & 0.68 & 0.71 & 0.68 \\
Fronting     & 0.66 & 0.67 & 0.68 & 0.67 & 0.67 & 0.71 & 0.67 & 0.69 & 0.68 \\
L\_FUNC\_T   & 0.67 & 0.66 & 0.75 & 0.66 & 0.67 & 0.65 & 0.67 & 0.65 & 0.67 \\
\bottomrule
\end{tabular}%
}
\caption{F1 score comparison between LD-score and the top-10 stylometric features 
(ranked by multiclass SHAP importance from \citet{Przystalski_2026}) trained on 
balanced subsets (per domain) of our benchmark and Ghostbuster dataset (125 human / 125 AI per domain, across 5 seeds). 
\textbf{Bold} denotes best per column.}
\label{tab:stylometry_snippet}
\end{table}



\section{Adversarial Results}
\label{app:adv}

Table~\ref{tab:auroc_adversarial} reports AUROC under three adversarial conditions. The augmented variants (LD-DNA, LD-Bino) outperform their base counterparts in 29 of 30 domain-attack comparisons, with the single exception being \texttt{medicine} under attack (B), where the difference is negligible ($-0.001$ and $-0.002$ respectively). Gains are largest under attack (B) (single letter removal), where Binoculars improves by up to $+0.029$ on \texttt{reddit\_eli5} and DNA improves by $+0.015$ on \texttt{finance}, suggesting the character-level LD-Score signal is particularly complementary when token distributions are locally perturbed. Attack (A) (paraphrase) yields the highest absolute AUROC values across all methods, with averages of $0.950$ and $0.944$ for DNA+LD and Bino+LD respectively, indicating that paraphrasing alone does not substantially degrade detection. Attack (C) (dual letter removal) is the hardest condition, producing the lowest overall AUROC, yet augmentation still improves both detectors consistently, with average gains of $+0.006$ and $+0.012$ for DNA and Binoculars. Across all attacks, \texttt{open\_qa} remains the most challenging domain, while \texttt{medicine} and \texttt{finance} are consistently the easiest.

\begin{table}[h]
\centering

\begin{tabular}{llccccccc}
\hline
\textbf{Attack} & \textbf{Dataset} & \textbf{DNA} & \textbf{LD-DNA} & \textbf{$\Delta$} & \textbf{Bino} & \textbf{LD-Bino} & \textbf{$\Delta$} \\

\hline
& finance      & $0.975_{\tiny\pm0.009}$ & $\textbf{0.979}_{\tiny\pm0.011}$ & $+0.004$ & $0.968_{\tiny\pm0.013}$ & $\textbf{0.978}_{\tiny\pm0.007}$ & $+0.010$ \\
& medicine     & $0.995_{\tiny\pm0.002}$ & $\textbf{0.996}_{\tiny\pm0.003}$ & $+0.001$ & $0.994_{\tiny\pm0.004}$ & $\textbf{0.995}_{\tiny\pm0.006}$ & $+0.001$ \\
\texttt{(A)} & open\_qa    & $0.834_{\tiny\pm0.012}$ & $\textbf{0.845}_{\tiny\pm0.025}$ & $+0.011$ & $0.808_{\tiny\pm0.011}$ & $\textbf{0.828}_{\tiny\pm0.028}$ & $+0.020$ \\
& reddit\_eli5 & $0.938_{\tiny\pm0.008}$ & $\textbf{0.948}_{\tiny\pm0.014}$ & $+0.010$ & $0.922_{\tiny\pm0.009}$ & $\textbf{0.937}_{\tiny\pm0.013}$ & $+0.015$ \\
& wiki\_csai   & $0.982_{\tiny\pm0.007}$ & $\textbf{0.984}_{\tiny\pm0.004}$ & $+0.002$ & $0.979_{\tiny\pm0.006}$ & $\textbf{0.983}_{\tiny\pm0.004}$ & $+0.004$ \\
& \textbf{avg} & $0.945_{\tiny\pm0.008}$ & $\textbf{0.950}_{\tiny\pm0.011}$ & $+0.005$ & $0.934_{\tiny\pm0.009}$ & $\textbf{0.944}_{\tiny\pm0.012}$ & $+0.010$ \\
\hline
& finance      & $0.946_{\tiny\pm0.005}$ & $\textbf{0.961}_{\tiny\pm0.005}$ & $+0.015$ & $0.916_{\tiny\pm0.014}$ & $\textbf{0.938}_{\tiny\pm0.013}$ & $+0.022$ \\
& medicine     & $\textbf{0.980}_{\tiny\pm0.005}$ & $0.979_{\tiny\pm0.003}$ & $-0.001$ & $\textbf{0.964}_{\tiny\pm0.007}$ & $0.962_{\tiny\pm0.006}$ & $-0.002$ \\
\texttt{(B)} & open\_qa & $0.756_{\tiny\pm0.042}$ & $\textbf{0.761}_{\tiny\pm0.026}$ & $+0.005$ & $0.719_{\tiny\pm0.042}$ & $\textbf{0.727}_{\tiny\pm0.032}$ & $+0.008$ \\
& reddit\_eli5 & $0.877_{\tiny\pm0.020}$ & $\textbf{0.886}_{\tiny\pm0.024}$ & $+0.009$ & $0.828_{\tiny\pm0.027}$ & $\textbf{0.857}_{\tiny\pm0.022}$ & $+0.029$ \\
& wiki\_csai   & $0.964_{\tiny\pm0.013}$ & $\textbf{0.968}_{\tiny\pm0.010}$ & $+0.004$ & $0.943_{\tiny\pm0.017}$ & $\textbf{0.954}_{\tiny\pm0.013}$ & $+0.011$ \\
& \textbf{avg} & $0.905_{\tiny\pm0.017}$ & $\textbf{0.911}_{\tiny\pm0.014}$ & $+0.006$ & $0.874_{\tiny\pm0.021}$ & $\textbf{0.888}_{\tiny\pm0.017}$ & $+0.014$ \\
\hline
& finance      & $0.938_{\tiny\pm0.014}$ & $\textbf{0.936}_{\tiny\pm0.013}$ & $-0.002$ & $0.898_{\tiny\pm0.016}$ & $\textbf{0.902}_{\tiny\pm0.018}$ & $+0.004$ \\
& medicine     & $0.959_{\tiny\pm0.011}$ & $\textbf{0.961}_{\tiny\pm0.010}$ & $+0.002$ & $0.943_{\tiny\pm0.009}$ & $\textbf{0.946}_{\tiny\pm0.012}$ & $+0.003$ \\
\texttt{(C)} & open\_qa & $0.712_{\tiny\pm0.020}$ & $\textbf{0.727}_{\tiny\pm0.034}$ & $+0.015$ & $0.674_{\tiny\pm0.023}$ & $\textbf{0.691}_{\tiny\pm0.033}$ & $+0.017$ \\
& reddit\_eli5 & $0.853_{\tiny\pm0.032}$ & $\textbf{0.863}_{\tiny\pm0.028}$ & $+0.010$ & $0.798_{\tiny\pm0.037}$ & $\textbf{0.819}_{\tiny\pm0.017}$ & $+0.021$ \\
& wiki\_csai   & $0.944_{\tiny\pm0.017}$ & $\textbf{0.946}_{\tiny\pm0.012}$ & $+0.002$ & $0.914_{\tiny\pm0.022}$ & $\textbf{0.928}_{\tiny\pm0.021}$ & $+0.014$ \\
& \textbf{avg} & $0.881_{\tiny\pm0.019}$ & $\textbf{0.887}_{\tiny\pm0.019}$ & $+0.006$ & $0.845_{\tiny\pm0.021}$ & $\textbf{0.857}_{\tiny\pm0.020}$ & $+0.012$ \\
\hline
\textbf{Avg} & & $0.910_{\tiny\pm0.014}$ & $\textbf{0.916}_{\tiny\pm0.015}$ & $+0.006$ & $0.885_{\tiny\pm0.017}$ & $\textbf{0.896}_{\tiny\pm0.016}$ & $+0.011$ \\
\hline
\end{tabular}
\caption{AUROC by Domain and Attack Type. (A): paraphrase attack, (B): single letter removal, (C): dual letter removal. \textbf{Bold} entries indicate which of the augmented or unaugmented variant performed better.}
\label{tab:auroc_adversarial}
\end{table}

Table~\ref{tab:f1_adversarial} reports F1 scores under the same three conditions. Unlike AUROC, augmentation yields mixed results: under attack (C) (dual letter removal), both LD-DNA and LD-Bino degrade on \texttt{open\_qa} ($-0.091$ and $-0.082$ respectively), and the overall averages are comparable or slightly worse than the base detectors. Attacks (A) and (B) tell a more favorable story, with consistent gains on \texttt{medicine}, \texttt{reddit\_eli5}, and \texttt{wiki\_csai}, and average improvements of $+0.008$ for both LD-DNA and LD-Bino under paraphrasing. As with AUROC, \texttt{open\_qa} remains the most adversarially vulnerable domain across all attacks and methods.

\begin{table}[h]
\centering

\begin{tabular}{llccccccc}
\hline
\textbf{Attack} & \textbf{Dataset} & \textbf{DNA} & \textbf{LD-DNA} & \textbf{$\Delta$} & \textbf{Bino} & \textbf{LD-Bino} & \textbf{$\Delta$} \\
\hline
& finance      & $\textbf{0.946}_{\tiny\pm0.012}$ & $0.946_{\tiny\pm0.011}$ & $+0.000$ & $0.929_{\tiny\pm0.016}$ & $\textbf{0.933}_{\tiny\pm0.012}$ & $+0.004$ \\
& medicine     & $0.973_{\tiny\pm0.009}$ & $\textbf{0.981}_{\tiny\pm0.008}$ & $+0.008$ & $0.968_{\tiny\pm0.016}$ & $\textbf{0.978}_{\tiny\pm0.011}$ & $+0.010$ \\
\texttt{(A)} & open\_qa    & $0.778_{\tiny\pm0.014}$ & $\textbf{0.785}_{\tiny\pm0.020}$ & $+0.007$ & $0.748_{\tiny\pm0.019}$ & $\textbf{0.750}_{\tiny\pm0.026}$ & $+0.002$ \\
& reddit\_eli5 & $0.869_{\tiny\pm0.010}$ & $\textbf{0.886}_{\tiny\pm0.016}$ & $+0.017$ & $0.841_{\tiny\pm0.015}$ & $\textbf{0.873}_{\tiny\pm0.010}$ & $+0.032$ \\
& wiki\_csai   & $0.936_{\tiny\pm0.021}$ & $\textbf{0.941}_{\tiny\pm0.018}$ & $+0.005$ & $\textbf{0.937}_{\tiny\pm0.019}$ & $0.933_{\tiny\pm0.011}$ & $-0.004$ \\
& \textbf{avg} & $0.900_{\tiny\pm0.013}$ & $\textbf{0.908}_{\tiny\pm0.015}$ & $+0.008$ & $0.885_{\tiny\pm0.017}$ & $\textbf{0.893}_{\tiny\pm0.014}$ & $+0.008$ \\

\hline
& finance      & $0.905_{\tiny\pm0.014}$ & $\textbf{0.909}_{\tiny\pm0.019}$ & $+0.004$ & $0.854_{\tiny\pm0.030}$ & $\textbf{0.864}_{\tiny\pm0.028}$ & $+0.010$ \\
& medicine     & $\textbf{0.924}_{\tiny\pm0.017}$ & $0.923_{\tiny\pm0.008}$ & $-0.001$ & $0.884_{\tiny\pm0.023}$ & $\textbf{0.891}_{\tiny\pm0.010}$ & $+0.007$ \\
\texttt{(B)} & open\_qa & $\textbf{0.720}_{\tiny\pm0.049}$ & $0.692_{\tiny\pm0.020}$ & $-0.028$ & $\textbf{0.694}_{\tiny\pm0.040}$ & $0.656_{\tiny\pm0.030}$ & $-0.038$ \\
& reddit\_eli5 & $0.781_{\tiny\pm0.023}$ & $\textbf{0.806}_{\tiny\pm0.014}$ & $+0.025$ & $0.740_{\tiny\pm0.033}$ & $\textbf{0.755}_{\tiny\pm0.025}$ & $+0.015$ \\
& wiki\_csai   & $0.907_{\tiny\pm0.023}$ & $\textbf{0.915}_{\tiny\pm0.025}$ & $+0.008$ & $0.878_{\tiny\pm0.016}$ & $\textbf{0.896}_{\tiny\pm0.018}$ & $+0.018$ \\
& \textbf{avg} & $0.847_{\tiny\pm0.025}$ & $\textbf{0.849}_{\tiny\pm0.017}$ & $+0.002$ & $0.810_{\tiny\pm0.028}$ & $\textbf{0.812}_{\tiny\pm0.022}$ & $+0.002$ \\

\hline
& finance      & $\textbf{0.880}_{\tiny\pm0.025}$ & $0.879_{\tiny\pm0.029}$ & $-0.001$ & $0.825_{\tiny\pm0.023}$ & $\textbf{0.841}_{\tiny\pm0.025}$ & $+0.016$ \\
& medicine     & $0.886_{\tiny\pm0.021}$ & $\textbf{0.898}_{\tiny\pm0.013}$ & $+0.012$ & $0.858_{\tiny\pm0.016}$ & $\textbf{0.876}_{\tiny\pm0.023}$ & $+0.018$ \\
\texttt{(C)} & open\_qa & $\textbf{0.692}_{\tiny\pm0.017}$ & $0.601_{\tiny\pm0.035}$ & $-0.091$ & $\textbf{0.667}_{\tiny\pm0.011}$ & $0.585_{\tiny\pm0.036}$ & $-0.082$ \\
& reddit\_eli5 & $0.774_{\tiny\pm0.023}$ & $\textbf{0.784}_{\tiny\pm0.015}$ & $+0.010$ & $\textbf{0.718}_{\tiny\pm0.027}$ & $0.718_{\tiny\pm0.018}$ & $+0.000$ \\
& wiki\_csai   & $0.875_{\tiny\pm0.014}$ & $\textbf{0.888}_{\tiny\pm0.024}$ & $+0.013$ & $0.837_{\tiny\pm0.027}$ & $\textbf{0.877}_{\tiny\pm0.021}$ & $+0.040$ \\
& \textbf{avg} & $\textbf{0.821}_{\tiny\pm0.020}$ & $0.810_{\tiny\pm0.023}$ & $-0.011$ & $\textbf{0.781}_{\tiny\pm0.021}$ & $0.779_{\tiny\pm0.025}$ & $-0.002$ \\
\hline
\textbf{Avg} & & $\textbf{0.856}_{\tiny\pm0.019}$ & $0.856_{\tiny\pm0.018}$ & $+0.000$ & $0.825_{\tiny\pm0.022}$ & $\textbf{0.828}_{\tiny\pm0.020}$ & $+0.003$ \\
\hline
\end{tabular}
\caption{F1 Score by Domain and Attack Type. (A): paraphrase attack, (B): single letter removal, (C): dual letter removal. \textbf{Bold} entries indicate which of the augmented or unaugmented variant performed better.}
\label{tab:f1_adversarial}
\end{table}

\end{document}